%% file: main.tex
\documentclass{article}

    \PassOptionsToPackage{numbers, compress}{natbib}
    \usepackage[final]{neurips_2022}

\PassOptionsToPackage{numbers, compress, comma}{natbib}

\usepackage[utf8]{inputenc} % allow utf-8 input
\usepackage[T1]{fontenc}    % use 8-bit T1 fonts
\usepackage{url}            % simple URL typesetting
\usepackage{booktabs}       % professional-quality tables
\usepackage{amsmath}
\usepackage{amssymb}
\usepackage{mathtools}
\usepackage{amsmath, bm, amsthm}      
\usepackage{nicefrac}       % compact symbols for 1/2, etc.
\usepackage{microtype}      % microtypography
\usepackage{graphicx,wrapfig}
\usepackage{subfig}
\usepackage[table,xcdraw]{xcolor}
\usepackage{tabularray}
\usepackage{hhline}
\usepackage{listings}

\usepackage{algorithm} 
\usepackage{algorithmic}
\usepackage{thmtools,thm-restate}
\definecolor{mydarkblue}{rgb}{0,0.08,0.45}

\usepackage[pagebackref=true,colorlinks,citecolor=mydarkblue,urlcolor=mydarkblue]{hyperref}
\renewcommand*\backref[1]{\ifx#1\relax\else(Cited on #1)\fi}
\usepackage{derivative}
\definecolor{headers}{RGB}{64,64,64}
\definecolor{cellcol}{RGB}{217,217,217}
\definecolor{first}{RGB}{198,247,198}
\definecolor{second}{RGB}{255,219,176}
\newcommand{\name}{$\Delta-$UQ~}
\newcolumntype{?}{!{\vrule width 2pt}}
\usepackage[export]{adjustbox}

\DeclareFixedFont{\ttb}{T1}{txtt}{bx}{n}{8} % for bold
\DeclareFixedFont{\ttm}{T1}{txtt}{m}{n}{8}  % for normal

\usepackage{color}
\definecolor{deepblue}{rgb}{0,0,0.5}
\definecolor{deepred}{rgb}{0.6,0,0}
\definecolor{deepgreen}{rgb}{0,0.5,0}

\usepackage{listings}

\newcommand\pythonstyle{\lstset{
language=Python,
basicstyle=\ttm,
morekeywords={self},              % Add keywords here
keywordstyle=\ttb\color{deepblue},
emph={MyClass,__init__},          % Custom highlighting
emphstyle=\ttb\color{deepred},    % Custom highlighting style
stringstyle=\color{deepgreen},
frame=tb,                         % Any extra options here
showstringspaces=false
}}

\lstnewenvironment{python}[1][]
{
\pythonstyle
\lstset{#1}
}
{}

\newcounter{numquote}
\newenvironment{lquote}{%
  \refstepcounter{numquote}%
  \quote}{\unskip~\fbox{\thenumquote}\endquote}

\title{Single Model Uncertainty Estimation via Stochastic Data Centering}

\author{%
 Jayaraman J. Thiagarajan\thanks{equal contribution} \\
  Lawrence Livermore National Laboratory\\
  \texttt{jjayaram@llnl.gov} \\
  \And
  Rushil Anirudh$^*$ \\
  Lawrence Livermore National Laboratory\\
  \texttt{anirudh1@llnl.gov} \\
  \AND
  Vivek Narayanaswamy \\
  Arizona State University\\
  \texttt{vnaray29@asu.edu} \\
  \And
  Peer-Timo Bremer \\
  Lawrence Livermore National Laboratory\\
  \texttt{bremer5@llnl.gov} \\
}

\begin{document}

\maketitle

\begin{abstract}

  We are interested in estimating the uncertainties of deep neural networks, which play an important role in many scientific and engineering problems. In this paper, we present a striking new finding that an ensemble of neural networks with the same weight initialization, trained on datasets that are shifted by a constant bias gives rise to slightly inconsistent trained models, where the differences in predictions are a strong indicator of epistemic uncertainties. Using the neural tangent kernel (NTK), we demonstrate that this phenomena occurs in part because the NTK is not shift-invariant. Since this is achieved via a trivial input transformation, we show that this behavior can therefore be approximated by training a single neural network -- using a technique that we call \name -- that estimates uncertainty around prediction by marginalizing out the effect of the biases during inference. We show that \name's uncertainty estimates are superior to many of the current methods on a variety of benchmarks-- outlier rejection, calibration under distribution shift, and sequential design optimization of black box functions. Code for \name can be accessed at \href{https://github.com/LLNL/DeltaUQ}{github.com/LLNL/DeltaUQ} 

\end{abstract}
\input{intro}
\input{background}

\input{methods}
\input{results}
\input{discussion}
\section*{Acknowledgement}
This work was performed under the auspices of the U.S. Department of Energy by Lawrence Livermore National Laboratory under Contract DE-AC52-07NA27344. Supported by the LDRD Program under projects 21-ERD-028, 22-ERD-006;  with IM release number LLNL-JRNL-836221.
\section*{Appendix}
The appendix contains the following additional information: (a) expanded derivations, (b) PyTorch code snippets for \name training \& inference; (c) results on calibration in standard regression settings on the UCI benchmarks dataset; (d) results on accuracy of predictions on ImageNet and CIFAR-10 and their corrupted variants; (e) calibration results on CIFAR10-C; (f) ablations on anchor-based training; (f) details on the design optimization experiment, including definitions for the benchmark functions and convergence plots.
\bibliography{refs,refs1}
\bibliographystyle{unsrtnat}
\input{checklist}

\clearpage
\appendix{\LARGE{\textbf{APPENDIX}}}
\input{supp_anchor_corruption.tex}
\input{supp_gan_expt}
\input{supp_design_opt}

\end{document}

%% file: intro.tex
\section{Introduction}
Accurately estimating uncertainties in a deep neural network (DNN) is an active area of research due to its implications in a wide range of scientific and engineering problems. Broadly, there are two kinds of uncertainties that are often considered -- (a) \emph{aleatoric}: uncertainties in the data generating process that are typically irreducible, and (b) \emph{epistemic}: uncertainties of the model that can be reduced by observing more data, which is the focus of our work. Some of the popular techniques for the latter include Bayesian methods \cite{wilson2020bayesian, he2020bayesian,neal2012bayesian,blundell2015weight} that use a prior on the network weights, Monte Carlo approximations such as MC Dropout \cite{gal2016dropout} that approximate sampling from the posterior, and empirical methods such as Deep Ensembles (DEns) \cite{lakshminarayanan2017simple}. In particular, DEns trains an ensemble of neural networks with different initializations, such that the uncertainty estimate on a test sample is given by the inconsistency between predictions from the member models. In practice, DEns has been found to often outperform other related methods~\cite{Ovadia2019,lakshminarayanan2017simple,van2020uncertainty}, but it comes at the cost of training several DNNs to obtain reliable uncertainties (typically $10-20$), which is a severe computational bottleneck when it comes to modern DNNs especially on large scale datasets. In light of this, there is increased interest in developing single model estimators that can still produce high quality uncertainties. There has been some promising work in this direction in the recent past, specific to deep classifiers \cite{van2020uncertainty} or regression models \cite{jain2021deup}, and has been shown to perform comparably to DEns in some use-cases.  

In this work, we first begin by exploring an alternate approach for constructing deep ensembles -- instead of using multiple randomized weight initializations, we propose to shift the training domain using random biases, such that every model in the ensemble is trained with data that has been shifted by a different \emph{constant} bias $\mathrm{c}$. Formally, for a labeled dataset $\{(\mathrm{x},y)\}$, the $k^{\text{th}}$ model is trained to fit $\{(\mathrm{x}-\mathrm{c}_k, y)\}$, where $\mathrm{c}_k$ (referred as an anchor) is of the same size as $\mathrm{x}$. Though this manipulation appears trivial, the kernel induced by deep networks is not inherently shift invariant \cite{jacot2018neural,tancik2020fourier}, thus implying each DNN learns a slightly different model due to the bias. This leads to one of our key observations:

% \begin{tabular}{?p{13cm}}
\begin{lquote}
\textbf{Anchor Ensemble:} \emph{When an ensemble of DNNs, with the same fixed initialization, is trained on a dataset shifted by random constant biases, the variation across the ensemble's predictions is a strong indicator of model uncertainty.}
 \label{quote:anchor_ensemble}
% \end{tabular}
\end{lquote}

Based on analysis with the neural tangent kernel (NTK) \cite{jacot2018neural}, we show that when the anchor $\mathrm{c}$ is made a random variable, the corresponding kernel is stochastic such that for each $\mathrm{c}$, the model converges to a slightly different NTK. Consequently, this anchoring-based ensembling provides a different approach to DEns for sampling solutions from the hypothesis space. However, interestingly, since the anchor ensembling emerges primarily from the act of a trivial data transformation, it lends itself to be approximated easily using a single DNN:  
\begin{lquote}
\textbf{\name}: \emph{For a randomly chosen anchor $\mathrm{c}$, \fbox{\ref{quote:anchor_ensemble}} can be approximated using a single DNN trained on the dataset transformed as $\{\mathrm{x},y\} \rightarrow \{[\mathrm{c}, \mathrm{x}-\mathrm{c}], y\}$.}
 \label{quote:deluq}
% \end{tabular}
\end{lquote}

% By expressing every sample as a tuple of a randomly chosen anchor, $c$ as $x_i \rightarrow [c,x_i-c] $, we can approximate anchor ensembles-like behavior by training a single DNN, where the choice of $c$ is randomized during training. 
During inference, we obtain multiple predictions for a given sample by varying the choice of the anchor, such that the standard deviation of the predictions is our estimate for uncertainty. Without affecting the performance of the model, we find that \name produces meaningful epistemic uncertainties that we validate in a variety of applications: outlier rejection,  calibration under distribution shifts on ImageNet, and sequential optimization of a large suite of black-box functions. We observe that \name consistently outperforms existing uncertainty estimates, while also being efficient to train as a single model estimator. With just a few simple changes, one can easily modify the training of any existing DNN to support \name estimation.
%We include a Torch implementation in the supplement, and will make our code public for easy benchmarking.

% 

% In this paper we propose a new single model uncertainty estimator, called \name, that addresses many of these concerns, while also being very effective on a variety of application benchmarks.

%% file: background.tex
\section{Background and Related Work}
Denote training data as $\mathcal{D} = \{(\mathrm{x}_i, y_i)\}_{i=1}^n$, where $\mathrm{x}_i \in \mathcal{X}$ and $y_i \in \mathcal{Y}$, to train a neural network $f(\boldsymbol{\theta}) \in \mathcal{H}$ with randomly initialized weights $\boldsymbol{\theta}_0$, such that a loss function $\mathcal{L}$ is minimized, \textit{i.e.}, $\arg \min_{\boldsymbol{\theta}} \mathcal{L}(f(\mathrm{x};\boldsymbol{\theta}), y)$. Here, $\mathcal{H}$ denotes the hypothesis space of potential solutions for fitting the observed data. Given a prior distribution on the weights $p(\boldsymbol{\theta})$, we can define the posterior over $\boldsymbol{\theta}$ as $p(\boldsymbol{\theta}|\mathcal{D})$ and subsequently, infer the posterior predictive distribution for a test sample $(\mathrm{x}_t, y_t)$. This can be used to quantify the uncertainty around the prediction as $p(y_t|\mathrm{x}_t, \mathcal{D}) = \int_{\boldsymbol{\theta}} p(y_t|\mathrm{x}_t, \boldsymbol{\theta})p(\boldsymbol{\theta}|\mathcal{D})d\boldsymbol{\theta}$. 

% \ptb{The agument chain in the next paragraph is weird. ... Estimating the posterio is difficult which has motivated ... ok ... the next step would be to explain why these Forare not working ... However, xyz ... Then you can talk about DEns .. Another approach is to use DEns which ....  Then you mention alternative solutions but never discuss why those are not ideal either}
A challenge for neural networks, however, is that the posterior $p(\boldsymbol{\theta}|\mathcal{D})$ is computationally intractable. This has motivated the use of \emph{Bayesian Neural Networks} (BNNs)~\cite{neal2012bayesian} using different approximations to the posterior including Monte-Carlo Dropout~\cite{gal2016dropout}, variational inference~\cite{graves2011practical, blundell2015weight}, and sampling methods such as Markov chain Monte Carlo~\cite{neal2012bayesian, welling2011bayesian}. In parallel, it has been empirically shown that Deep Ensembles (DEns) \cite{lakshminarayanan2017simple} often tend to outperform Bayesian methods in terms of model calibration performance, even under challenging distribution shifts~\cite{Ovadia2019}.  
%Conceptually, DEns involves training multiple models with different initializations, such that the prediction at test time is obtained as the average of predictions from each of the members in the ensemble, and the corresponding variance is a strong indicator of the epistemic uncertainty. 
The success of DEns has been attributed to its ability to sample different functional modes \cite{fort2019deep} from the hypothesis space, and thus approximate the posterior predictive distribution~\cite{wilson2020bayesian}. However, a critical limitation of DEns is the need to train multiple models (typically $10-20$) in order to obtain well calibrated uncertainties, which can be impractical for complex model architectures that have become commonplace today. 
%This has motivated alternative solutions that can reliably approximate the uncertainty around a prediction with just a single model~\cite{van2020uncertainty, jain2021deup}.
% \ptb{There is a very abrupt shift here between paragraphs. It is not clear how the above discussion leads to the one below}

Characterizing the behavior of deep uncertainty estimators has mostly been done using empirical evaluation based on model calibration or out-of-distribution detection, but the recent advances in the neural tangent kernel (NTK) theory \cite{jacot2018neural, arora2019fine, bietti2019inductive, lee2019wide} provide a convenient framework for more rigorous analysis. The basic idea of NTK is that, when the width of a neural network tends to infinity and the learning rate of stochastic gradient descent (SGD) tends to zero, the function $f(\mathrm{x};\boldsymbol{\theta})$ converges to a solution obtained by kernel regression using the NTK defined as
% \begin{equation}
    % \label{eqn:ntk}
    $\mathbf{K}_{\mathrm{x}_i\mathrm{x}_j} = \mathbb{E}_{\boldsymbol{\theta}}\left\langle \pdv{f(\mathrm{x}_i,\boldsymbol{\theta})}{\boldsymbol{\theta}}, \pdv{f(\mathrm{x}_j,\boldsymbol{\theta})}{\boldsymbol{\theta}} \right\rangle.$
% \end{equation}
When the samples $\mathrm{x}_i, \mathrm{x}_j \in \mathcal{S}^{d-1}$, i.e.,  points on the hypersphere and have unit norm, the NTK for a simple 2 layer ReLU MLP can be simplified as a dot product kernel \cite{arora2019fine,bietti2019inductive,lee2019wide}:
\begin{equation}
    \label{eqn:mlp_ntk}
\mathbf{K}_{\mathrm{x}_i\mathrm{x}_j} =     h_{\text{NTK}}(\mathrm{x}_i^\top \mathrm{x}_j) = \frac{1}{2\pi}\mathrm{x}_i^\top \mathrm{x}_j(\pi - \cos^{-1}(\mathrm{x}_i^\top \mathrm{x}_j))
\end{equation}The NTK framework and its extensions that enable a posterior interpretation in the infinite limit have been used to study deep ensembles~\cite{he2020bayesian}. Previous work \cite{lee2018deep,gMatthews2018gaussian,novak2019bayesian} has also shown the existence of a distinct yet related kernel, referred to as the neural network Gaussian process (NNGP) kernel, where the initialization tends to a GP in infinite width limit.  
% \ptb{This sounds off .... how can the same network converge to different kernels. YOu say somehting about initialization? I feel like there is some information missing here that I do not have}

%% file: methods.tex
\section{Uncertainty estimation with \name}

As discussed above, most existing frameworks sample the hypothesis space either through different random initializations of ($\boldsymbol{\theta_0}$), perturbing the weight space after training, or using Bayesian methods by placing a prior on $p(\boldsymbol{\theta})$. Here, we propose to use a new strategy that involves injecting multiple trivial biases into the training data and analyze the resulting models using the NTK framework. 

\subsection{Anchor Ensembles: ensembling by injecting trivial biases}
Let us examine the scenario where we shift an entire dataset (both train and validation) using a constant bias, $\mathrm{c}$, to obtain a new dataset $\mathcal{D}_{\mathrm{c}}$ using which we train the model $f_{\mathrm{c}}$. Since we always choose $\mathrm{c}$ from the training distribution at random, this has the effect of zero-centering the dataset around different training points. Let $\{f_{\mathrm{c}_1}, f_{\mathrm{c}_2}, \dots, f_{\mathrm{c}_k}\}$ denote the set of models trained using different $\mathrm{c}$'s, then our goal is to characterize the relationship between them as a function of $\mathrm{c}$. If the NTK induced by $f$ is shift-invariant (for e.g., when Fourier features~\cite{tancik2020fourier} are used), the shifts will make no difference, resulting in identical models $f_{\mathrm{c}_1} = \dots=f_{\mathrm{c}_k}$. However, since NTKs for models like MLPs and CNNs are not inherently shift-invariant \cite{lee2019wide}\footnote{By shift-invariance, we refer to the \emph{domain} shift induced because of the random bias, as opposed to \emph{spatial} shift-invariance that is more commonly associated with convolutional operators.}, we find that the models lead to an effective deep ensemble, wherein the variation across the predictions is a strong indicator of epistemic uncertainties. 

\textbf{Effect of shifted training on NTK:} We are interested in understanding how \eqref{eqn:mlp_ntk} changes when the \emph{entire training domain} is shifted by $\mathrm{c}$ -- i.e., $h_{\text{NTK}}((\mathrm{x}_i-\mathrm{c})^\top (\mathrm{x}_j-\mathrm{c}))$. Without loss of generality, for the sake of notational convenience, we assume $\mathrm{x}_i-\mathrm{c}$ and $\mathrm{x}_j-\mathrm{c}$ are also made unit norm. To simplify the expansion, we use a Taylor series expansion for the $\cos^{-1}$ function: $\cos^{-1}(\mathrm{u}-\mathrm{c}) \approx \cos^{-1}(\mathrm{u}) + \frac{\mathrm{c}}{\sqrt{1-(\mathrm{u}-\mathrm{c})^2}}$. 

\begin{figure}[!t]
    \centering
    \includegraphics[width = 0.99\linewidth, trim=0cm 0cm 0cm 0cm,clip]{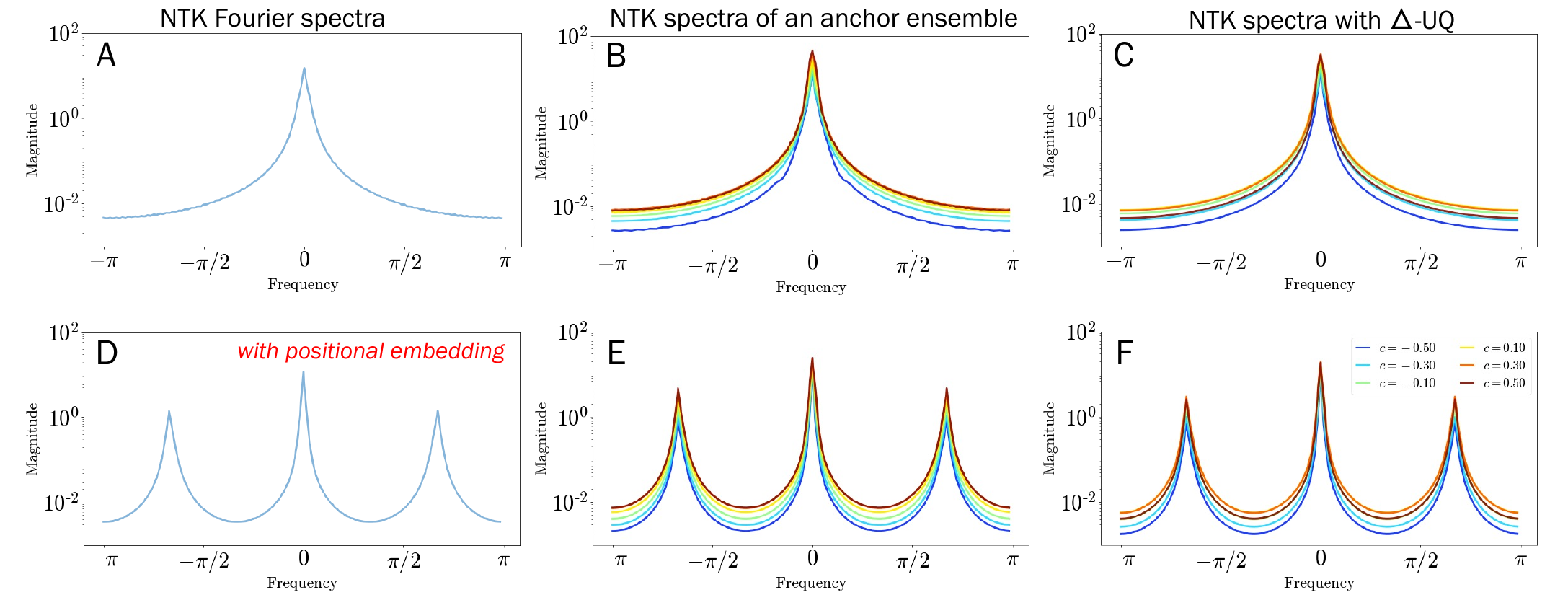}
    \caption{\small{Fourier spectrum of an NTK for an MLP model (A,D); spectra of an anchor ensemble (B, E); and NTK spectra using \name (C, F). Bottom row shows NTK spectra when inputs are passed through a sinusoidal PE. We make two key observations -- a) trivial shifts in the input domain cause the \emph{effective} NTK to be distinct as a function of the shift $c$, as seen in eqn. \eqref{eqn:shifted_ntk}; and b) \name achieves a similar effect but with a single model.}}
    \label{fig:ntk_spectral}
\end{figure}

Expanding $(\mathrm{x}_i-\mathrm{c})^\top (\mathrm{x}_j-\mathrm{c})$ as $\mathrm{x}_i^\top \mathrm{x}_j - \mathrm{c}^\top(\mathrm{x}_i + \mathrm{x}_j - \mathrm{c})$ and letting $\mathrm{v} = (\mathrm{x}_i + \mathrm{x}_j - \mathrm{c})$, we obtain the expression for $h_{\text{NTK}}$ under a shifted domain as follows:
\begin{flalign}
\label{eqn:shifted_ntk}
    & \mathbf{K}_{(\mathrm{x}_i-\mathrm{c})(\mathrm{x}_j-\mathrm{c})} =  \frac{1}{2\pi}(\mathrm{x}_i^\top \mathrm{x}_j - \mathrm{c}^\top \mathrm{v})(\pi - \cos^{-1}(\mathrm{x}_i^\top \mathrm{x}_j - \mathrm{c}^\top \mathrm{v})) \notag &\\
    & \approx \frac{1}{2\pi}\mathrm{x}_i^\top \mathrm{x}_j(\pi - \cos^{-1}(\mathrm{x}_i^\top \mathrm{x}_j)) - \frac{1}{2\pi}\mathrm{c}^\top \mathrm{v}(\pi - \cos^{-1}(\mathrm{x}_i^\top \mathrm{x}_j)) - \frac{\mathrm{c}(\mathrm{x}_i^\top \mathrm{x}_j - \mathrm{c}^\top \mathrm{v})}{2\pi \sqrt{1 - (\mathrm{x}_i^\top \mathrm{x}_j - \mathrm{c}^\top \mathrm{v})^2}}\notag&\\
    & = \mathbf{K}_{\mathrm{x}_i\mathrm{x}_j} - \Gamma_{\mathrm{x}_i, \mathrm{x}_j, \mathrm{c}},
\end{flalign}
where we combine all terms dependent on $c$ into $\Gamma_{\mathrm{x}_i, \mathrm{x}_j, \mathrm{c}}$, which also behaves as a dot product kernel. From \eqref{eqn:shifted_ntk}, we note that a trivial shift in the domain results in a non-trivial shift in the NTK function itself. In other words, \eqref{eqn:shifted_ntk} outlines the \emph{effective} NTK as a function of $\mathrm{c}$. We also note that $\mathbf{\Gamma}$ does not affect the spectral properties of the original NTK, as we observe in Figure \ref{fig:ntk_spectral}.

% We expand our intuition based on NTK for neural networks \cite{jacot2018neural}. Lee et al. \cite{lee2019wide} study the linearized regime of an NN using the Taylor expansion given a randomly initialized neural network $f(\boldsymbol{\theta}_0)$:
% \begin{equation}
% \label{eq:ntk_linear}
%     f_t^{\text{lin}}(x) = f_0(x) + \nabla_\boldsymbol{\theta} f(x, \boldsymbol{\theta}_0) \Delta \boldsymbol{\theta}_t 
% \end{equation}
Let us now consider the prediction on a test sample $\mathrm{x}_t$ in the limit as the inner layer widths grow to infinity. It has been shown that (c.f. \cite{lee2019wide,bietti2019inductive}): 
\begin{equation}
\label{eqn:ntk_limit}
\centering
f_\infty(\mathrm{x}_t) = f_0(\mathrm{x}_t) - \mathbf{K}_{\mathrm{x}_t\mathbf{X}}\mathbf{K}_{\mathbf{X} \mathbf{X}}^{-1}(f_0(\mathbf{X})-\mathbf{Y}),
\end{equation}where $\mathbf{X}$ is the matrix of all training data samples. As before, we consider the case where the domain is shifted by $\mathrm{c}$. Using \eqref{eqn:ntk_limit}:
\begin{align}
    &f_\infty(\mathrm{x}_t-\mathrm{c}) = f_0(\mathrm{x}_t-\mathrm{c}) - \mathbf{K}_{(\mathrm{x}_t-\mathrm{c})(\mathbf{X}-\mathrm{c})}\mathbf{K}_{(\mathbf{X}-\mathrm{c}) (\mathbf{X}-\mathrm{c})}^{-1}(f_0(\mathbf{X}-\mathrm{c})-\mathbf{Y}) \notag &\\
    &\approx f_0(\mathrm{x}_t-\mathrm{c}) - (\mathbf{K}_{\mathrm{x}_t\mathbf{X}} - \Gamma_{\mathrm{x}_t, \mathbf{X},\mathrm{c}})(\mathbf{K}_{\mathbf{X}\mathbf{X}} - \Gamma_{\mathbf{X}, \mathbf{X},\mathrm{c}})^{-1}(f_0(\mathbf{X}-\mathrm{c})-\mathbf{Y})\notag &\\
     &= f_0(\mathrm{x}_t-\mathrm{c}) - (\mathbf{K}_{\mathrm{x}_t\mathbf{X}} - \Gamma_{\mathrm{x}_t, \mathbf{X},\mathrm{c}})\left(\mathbf{K}_{\mathbf{X}\mathbf{X}}^{-1} + \sum_{m=1}^\infty (\mathbf{K}_{\mathbf{X}\mathbf{X}}^{-1}\Gamma_{\mathbf{X}, \mathbf{X},\mathrm{c}})^m\mathbf{K}_{\mathbf{X}\mathbf{X}}^{-1}\right)(f_0(\mathbf{X}-\mathrm{c})-\mathbf{Y})\label{eqn:woodbury}&\\
    &\approx \underbrace{f_0(\mathrm{x}_t) - \mathbf{K}_{\mathrm{x}_t\mathbf{X}}\mathbf{K}_{X\mathbf{X}}^{-1}(f_0(\mathbf{X}) - \mathbf{Y})}_{\text{deterministic for fixed }\boldsymbol{\theta}_0} - \underbrace{g(\mathrm{c}, \mathrm{x}_t, \mathbf{X}, \mathbf{Y})}_{\text{random due to }\mathrm{c}}\label{eqn:ntk_anchor_pred}
    %  & = f_\infty(x) - g(x, \mathbf{X}, \mathcal{Y})
\end{align}
In \eqref{eqn:woodbury}, we utilize Woodbury's Inverse Identity \cite{woodbury1950inverting} to expand the inverse of a sum of matrices. Next, in \eqref{eqn:ntk_anchor_pred}, we combine all the terms dependent on $\mathrm{c}$ into a function $g$. Note, we also expand $f_0(\mathrm{x}-\mathrm{c})$ using the Taylor series approximation to represent it as a sum of $f_0(\mathrm{x})$ and other terms (details of this derivation can be found in the appendix). 

% We can make a few observations regarding \eqref{eqn:ntk_anchor_pred}. 

For a given initialization $\boldsymbol{\theta}_0$, the first term is deterministic, and exactly the same as \eqref{eqn:ntk_limit}, since all other the terms are fixed. However, we see that the second term can vary based on the choice of $\mathrm{c}$. Existing ensembling approaches~\cite{lakshminarayanan2017simple} rely on the randomness of the initialization $\boldsymbol{\theta}_0$ to pick diverse solutions from the hypothesis space~\cite{he2020bayesian}. In contrast, equations \eqref{eqn:shifted_ntk} and \eqref{eqn:ntk_anchor_pred} suggest that even for a fixed $\boldsymbol{\theta}_0$, it is possible to make the NTK stochastic (in $\mathrm{c}$) by shifting the entire input domain. To better understand how the NTK actually changes from these expressions, we study the spectral properties of a simple MLP network empirically, following the analysis in ~\cite{tancik2020fourier}.

\textbf{Spectral properties of shifted NTKs.}
We compute the Fourier spectra using the same MLP on several shifted domains in Figure \ref{fig:ntk_spectral}(B). The original spectrum for the MLP without any shift in the training domain is shown for comparison in \ref{fig:ntk_spectral}(A). As indicated by \eqref{eqn:shifted_ntk}, we see that each individual shift leads to a different NTK (as indicated by the spectra), either by flattening it or making it narrower in the frequency domain. The same behavior persists even when we construct positional embeddings (PE), based on sinusoidal functions, prior to building the MLP model (\ref{fig:ntk_spectral}(D-E)). This leads to one of our main findings stated in \fbox{\ref{quote:anchor_ensemble}}, and illustrated in Figure \ref{fig:anchor_example}.

% \begin{tabular}{?p{13cm}}
% \emph{An anchoring-based ensemble of neural networks with the same fixed initialization, trained on shifted variations of a dataset produces meaningful epistemic uncertainties.}
% \end{tabular}

%Note, all terms in \eqref{eqn:ntk_anchor} here are affected by the anchor unlike in deep ensembles \cite{lakshminarayanan2017simple}, where the randomness comes primarily from the choice of $f_0(x)$, including in Bayesian deep ensembles \cite{he2020bayesian}.
%%
\subsection{\name: rolling anchor ensembles into a single model}
\label{sec:deltauq}
Since different models in an anchoring-based ensemble are trained with the same initialization, we present a new technique to approximate the epistemic uncertainties using a single neural network. More specifically, we perform a simple coordinate transformation by lifting the domain to a higher dimension as $\mathcal{E}: \mathrm{x} \rightarrow \{\mathrm{c}, \mathrm{x}-\mathrm{c}\}$, we refer to the residual by $\Delta = \mathrm{x}-\mathrm{c}$. This transformation allows the use of multiple representations (w.r.t. different anchors) for the same input sample $\mathrm{x}$, \textit{i.e.}, $f_{\Delta}(\{\mathrm{c}_1, \mathrm{x}-\mathrm{c}_1\}) = f_{\Delta}(\{\mathrm{c}_2, \mathrm{x}-\mathrm{c}_2\}) = \dots = f_{\Delta}(\{\mathrm{c}_k, \mathrm{x}-\mathrm{c}_k\})$, where $f_{\Delta}$ refers to the \name model that takes the tuple $(\{\mathrm{c}_k, \mathrm{x}-\mathrm{c}_k\})$ and predicts the target $y$.

It is easy to see that $\left[\mathrm{c},\mathrm{x}_i-\mathrm{c}\right]^\top\left[\mathrm{c},\mathrm{x}_j-\mathrm{c}\right] = \mathrm{x}_i^\top \mathrm{x}_j - \mathrm{c}^\top(\mathrm{x}_i + \mathrm{x}_j - 2\mathrm{c})$.
% \begin{align}
% \label{eq:deluq_dot}
% (\mathrm{x}_i-c)^\top (\mathrm{x}_j-c) & = \mathrm{x}_i^\top \mathrm{x}_j - \mathrm{c}^\top(\mathrm{x}_i + \mathrm{x}_j - \mathrm{c})\notag\\ \left[\mathrm{c},\mathrm{x}_i-\mathrm{c}\right]^\top\left[\mathrm{c},\mathrm{x}_j-\mathrm{c}\right] & = \mathrm{x}_i^\top \mathrm{x}_j - \mathrm{c}^\top(\mathrm{x}_i + \mathrm{x}_j - 2\mathrm{c})
% \end{align}
That is, the dot product of the transformed inputs takes the same form as before (except for a scaling factor). Therefore, the expressions for the equivalent NTK for different anchor shifts, seen in \eqref{eqn:shifted_ntk}, and the corresponding prediction on a test sample seen in \eqref{eqn:ntk_anchor_pred} remain the same for \name, by setting $\mathrm{v} = \mathrm{x}_i + \mathrm{x}_j - 2\mathrm{c}$. A similar form is also obtained in the more general case when different anchors are used for different samples (as shown in the appendix), which leads to our main claim stated in \fbox{\ref{quote:deluq}}, that the \name model achieves similar perturbations of the NTK as an anchor ensemble, based on the choice of $\mathrm{c}$.
 
\textbf{Training.} During training, for every input $\mathrm{x}_i$ we choose an anchor as a random sample from the training dataset. Subsequently, we obtain the coordinate transformation $\{[\mathrm{c}, \mathrm{x}_i-c],y_i\}$, using which we train the model. With vector-valued data, this is implemented as a simple concatenation. In the case of image data, we append the channels to create a $6-$dimensional tensor (for a 3-channel  RGB image). We show simple a PyTorch snippet for training \name in Figure \ref{fig:pseudocode}. Other than increasing the number of parameters in the first layer of the network, $\Delta-$UQ does not incur additional computational overheads. 

% \begin{algorithm}[htb!]
%   \caption{PyTorch-style pseudo-code for \name.}
%   \label{alg:deltauq}
   
%     \definecolor{codeblue}{rgb}{0.25,0.5,0.5}
%     \lstset{
%       basicstyle=\fontsize{7.2pt}{7.2pt}\ttfamily\bfseries,
%       commentstyle=\fontsize{7.2pt}{7.2pt}\color{codeblue},
%       keywordstyle=\fontsize{7.2pt}{7.2pt},
%     }
% \begin{lstlisting}[language=python]
\begin{wrapfigure}{r}{6cm}
\vspace{-15pt}
\caption{\small{Mini-batch training with \name.}}
\begin{python}
for inputs, targets in trainloader:
    A = Shuffle(inputs) %% Anchors
    D = inputs-A %% Delta
    X_d = torch.cat([A, D],axis=1)
    y_d = model(X_d) %% prediction
    loss = criterion(y_d,targets)
\end{python}

\label{fig:pseudocode}
\vspace{-10pt}
\end{wrapfigure}

\begin{figure}[!t]
    \centering
    \includegraphics[width = 0.85\linewidth]{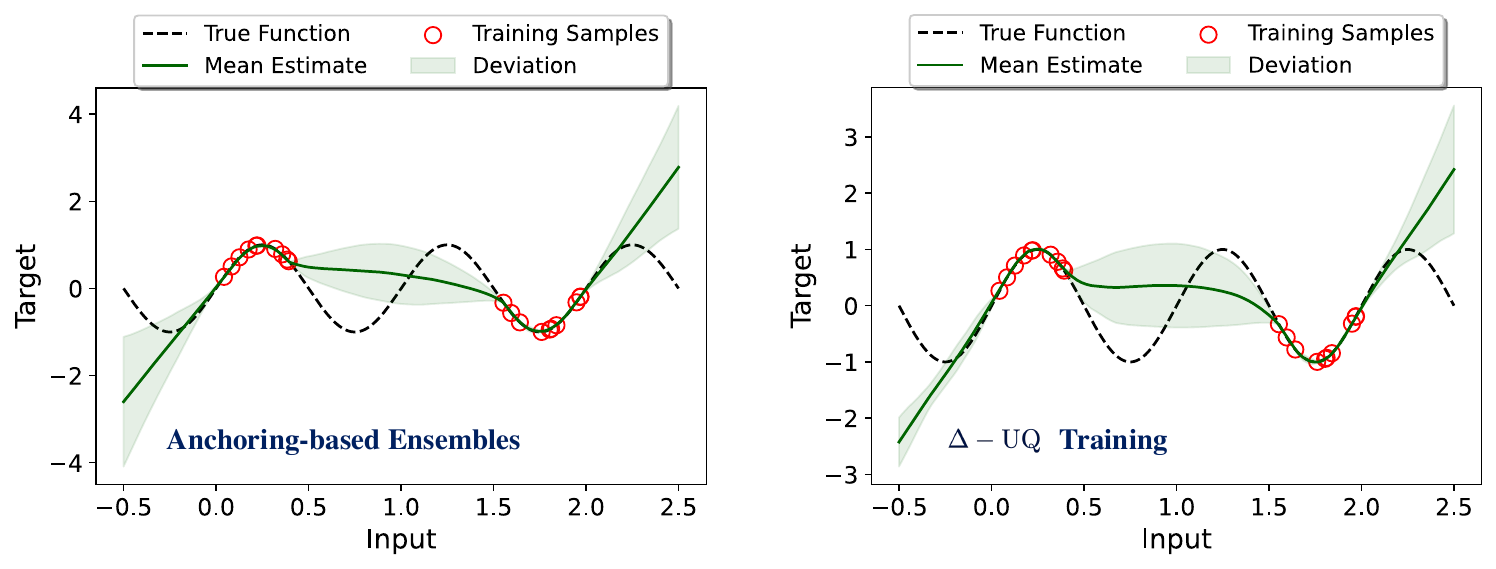}
    \vspace{-10pt}
    \caption{\small{Comparing anchor ensembles and \name in function fitting with an MLP. As expected, we see that the disagreement between models in an anchor ensemble correlate strongly with the epistemic uncertainty, and that \name, with a single model, matches this behavior very closely.}}
    \label{fig:anchor_example}
    \vspace{-10pt}
\end{figure}

Over the course of training, in expectation, every training pair gets combined with a large number of anchors. Since the prediction on this training pair -- regardless of anchor choice -- must always be the same, this places an implicit consistency in the predictions that they must be similar no matter which anchor is chosen. This consistency trades-off with diversity of the kinds of functions that can be learned when compared with an anchor ensemble, where the models are trained independently. This can be seen in the comparisons of the NTK spectra for \name with anchor ensembles in Figures \ref{fig:ntk_spectral}(C) and (F). In practice, however, we find that the diversity from this single model is still sufficiently large, to estimate good quality uncertainties.

\textbf{Inference.} For a test sample $\mathrm{x}_t$, we obtain the prediction from \name as the mean prediction across several randomly chosen anchors; and the standard deviation around these predictions is our estimate for the epistemic uncertainty. In other words, we marginalize out the effect of anchors to obtain the final prediction mean and uncertainty. Formally, the predictive distribution is given by $p(y_t|\mathrm{x}_t) = \int_{\mathrm{c}\in\mathbf{X}}p(y_t|\mathrm{x}_t, \mathrm{c}, \boldsymbol{\theta})p(\mathrm{c}) d\mathrm{c}$. In practice, for a trained \name model specified as $\boldsymbol{\theta}^*$, we compute the sample mean and uncertainty around it as: 
\begin{equation}
    \label{eqn:deluq_pred}
    \boldsymbol\mu(y_t|\mathrm{x}_t) = \frac{1}{K}\sum_{k=1}^K f([\mathrm{c}_k,\mathrm{x}_t-\mathrm{c}_k], \boldsymbol{\theta}^*); ~~ \boldsymbol\sigma(y_t|\mathrm{x}_t) = \sqrt{\frac{1}{K-1}\sum_{k=1}^K (f([\mathrm{c}_k,\mathrm{x}_t-\mathrm{c}_k], \boldsymbol{\theta}^*) - \boldsymbol\mu)^2 },
\end{equation}
% The \name model's predictions vary depending on the choice of the anchor, which act as a conditioning variable. This is reflected in the \emph{effective} NTK for each anchor as well, as seen in Figure \ref{fig:ntk_spectral}(c). The kernel spectrum is adjusted based on the choice of the anchor, indicating that multiple different kernels can be realized by simply choosing a different anchor. Consequently, the disagreement across predictions from multiple anchors is reflective of epistemic uncertainty around the test sample. 

\paragraph{Discussion.} In Figure \ref{fig:anchor_example}, we show a $1$D regression example using $20$ training examples along with the predicted mean and estimated uncertainties. As it can be seen, both the anchoring-based ensemble (left) and \name training show higher epistemic uncertainties around regions with no training samples. In the ensembles version, we train $20$ different networks -- while in \name we use just a single network, where the uncertainty is obtained using all $20$ anchors during inference.

We always use only a single anchor for an input during every training iteration, but multiple anchors during inference time. In theory, multiple anchors could be used during training as well, where the loss is imposed on the mean (obtained with multiple anchors). However, we find that simply using a single random anchor in each iteration achieves a similar effect, as it enforces a consistency that the same training pair $(\mathrm{x},y)$ when combined with many different anchors over the course of training as $[\mathrm{c}_1, \mathrm{x}_i-\mathrm{c}_1], [\mathrm{c}_2, \mathrm{x}_i-\mathrm{c}_2], \dots, [\mathrm{c}_k, \mathrm{x}_i-\mathrm{c}_k]$ must all produce the same prediction, $y$. 
%When a typical model trained for $T$ epochs, every training pair gets exposed to $T$ different anchors on average.

\name relies on randomly drawn anchors for uncertainty estimation, which is similar to DEns \cite{lakshminarayanan2017simple}, that relies on the diversity of the base learners in an ensemble. Arguably, sampling a random set of anchors from the training distribution is simpler than sampling from the posterior $p(\boldsymbol{\theta}|\mathcal{D})$. Furthermore, since every anchor realizes a slightly different function, using a small number of anchors ($10-20$) during inference is typically sufficient to obtain high quality estimates as we show in our experiments. Finally, due to the nature of training with random anchors we also see that \name produces particularly effective uncertainty estimates when the training set size is small, and this proves to be very useful in applications such as sequential optimization.
%
% Anchors are central to \name's uncertainty estimates since they directly control the kinds of perturbations induced into the networks fit (see \eqref{eqn:shifted_ntk}). For 
%
% small sample sizes, small number of anchors, but also scales to large datasets
%%% Local Variables:
%%% mode: latex
%%% TeX-master: "main"
%%% End:

%% file: results.tex
\section{Experiments}
We validate our approach in this section using a variety of applications and benchmarks -- (a) first, we consider the utility of epistemic uncertainties in object recognition problems where they have been successfully used for outlier rejection and calibrating models under distribution shifts. We show that \name can be very effective even with large-scale datasets like ImageNet \cite{russakovsky2015imagenet}; (b) next, we consider the challenging problem of sequential design optimization of black-box functions, where the goal is to maximize a scalar function of interest with the fewest number of sample evaluations. Using a Bayesian optimization setup, we show that the uncertainties obtained using \name outperform many competitive methods across an extensive suite of black-box functions.

\paragraph{a. Outlier rejection.}
A popular application for epistemic uncertainties is in rejecting outliers since, by definition, they are in regions outside of the training distribution. As such, we expect the model to produce highly uncertain predictions for these images, which should help us design an effective OOD detector. To evaluate this hypothesis, we train a modified ResNet-50 \cite{he2016deep} model on ImageNet that accepts $6$ input channels (anchor, $\Delta$) as outlined earlier. We train the model using standard hyperparameter settings except, training it longer for 120 epochs -- our top-1 accuracy ($76.1$) matches that of a standard ResNet-50. Specifically for images, we found that corrupting the anchors with common transforms like random crops, Gaussian blurs, and color jitter improves performance. That is, instead of $[c, x-c]$, we use $[\mathcal{T}(c), x-c]$ where $\mathcal{T}$ is a simple transform like Gaussian blur. The uncertainty for a test sample is given by the standard deviation of the logits obtained by varying the anchors. To obtain a scalar statistic, we compute the mean across all classes as the uncertainty score for that sample. We provide more details on the transformations used for corrupting the anchor, and the accuracies of our model across shifted variants in more detail in the appendix.
\begin{figure}[!htb]
\centering
\subfloat[\small{Uncertainties change meaningfully as outliers get more severe}]{\includegraphics[width=0.4\linewidth,valign=c]{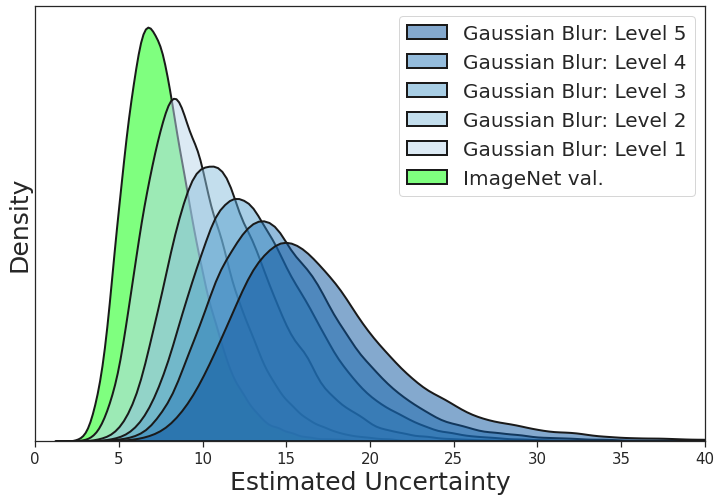}
    \label{fig:imagenet_density}}
\subfloat[\small{Uncertainties from \name for outlier rejection}]{
\vspace{0pt}
\small{
\begin{tabular}[c]{p{0.8in}|p{0.4in}p{0.4in}c}
	\toprule
    Method & AUROC~$\uparrow$& DTACC~$\uparrow$& AUPR-in/out~$\uparrow$\\ 
    \midrule
	ResNet-50 \cite{he2016deep} & 93.36 & 86.08 & 92.82 / 93.71 \\
	Temp-Scal \cite{guo2017calibration} & 93.71 & 86.47& 93.21 / 94.01 \\
	Deep Ens \cite{lakshminarayanan2017simple} & 95.49 & 88.82 & 95.31 / 95.64\\
	MC Dropout \cite{gal2016dropout} & 96.38 & 89.98 & 96.16 / 96.67\\
	SVI \cite{blundell2015weight} & 96.40 & 90.03 & 95.97 / 96.83\\
% 	SVI-AvUC \cite{krishnan2020improving} & 97.60 & 92.07 & 97.39 / 97.85\\
	\midrule
	\name (ours)&  \textbf{97.49} & \textbf{91.90} &  \textbf{97.56} /  \textbf{97.47}\\
	\bottomrule
\end{tabular}}
\label{tab:Imagenet}}
\caption{\small{\textbf{Rejecting outliers with epistemic uncertainties:} We evaluate \name on the benchmark introduced by \cite{krishnan2020improving} where we use Gaussian Blur of level 5 intensity as the outliers from the ImageNet validation set. At inference, uncertainties are estimated as the mean of std. dev of predictions obtained with $10$ anchors.}}
 \label{fig:imagenet_ood}
 \vspace{-5pt}
\end{figure}

We show the results for outlier rejection in \ref{tab:Imagenet}, where we follow the protocol established in \cite{krishnan2020improving}, that uses a Gaussian blur of intensity 5 from ImageNet-C \cite{hendrycks2018benchmarking} as the outlier set, and the clean ImageNet validation data as inliers. We use the estimated uncertainty obtained with $10$ anchors as in \eqref{eqn:deluq_pred} as our score for outlier rejection and report commonly used metrics such as AUROC, Detection Accuracy (DTACC), and AUPR-in/out. We note that, just the inconsistency of predictions obtained using \name outperforms many baselines including mean-field stochastic variational inference (SVI) \cite{graves2011practical, blundell2015weight}, SVI-AvUC \cite{krishnan2020improving}, Monte Carlo dropout \cite{gal2016dropout}, and temp. scaling \cite{guo2017calibration}. While this can be further improved by taking the mean prediction into account, similar to existing approaches for semantic novelty detection (with scores such as entropy, energy \cite{energyood} etc.), our focus here is to evaluate the quality of uncertainty alone.  In Figure \ref{fig:imagenet_density}, we observe that \name's uncertainty estimate changes smoothly as the outliers become farther away (more severe intensity) from the training distribution.
\begin{table}[!t]
\caption{\small{\textbf{Calibration under distribution shift:} A ResNet-50 model that is tempered by uncertainties obtained from \name (see text) outperforms several competitive baselines averaged across 16 different corruptions of ImageNet-C at highest severity level 5.}}
\label{tab:Imagenet_calibration}
\small{\begin{tblr}{
colspec ={X[1] X[1.8] X[1] X[1] X[1] X[1] X[1] X[1] X[1] X[1]},
hlines = {black,0pt},
vlines = {black,0.5pt},
cell{1}{1} = {r=2, c=1}{c,headers,fg=white},
cell{1}{2} = {r=2, c=1}{c,headers,fg=white},
cell{1}{3} = {r=2, c=1}{c,headers,fg=white},
cell{1}{4} = {r=2, c=1}{c,headers,fg=white},
cell{1}{5} = {r=2, c=1}{c,headers,fg=white},
cell{1}{6} = {r=2, c=1}{c,headers,fg=white},
cell{1}{7} = {r=2, c=1}{c,headers,fg=white},
cell{1}{8} = {r=2, c=1}{c,headers,fg=white},
cell{1}{9} = {r=2, c=1}{c,headers,fg=white},
cell{1}{10} = {r=2, c=1}{c,headers,fg=white},
row{3-14}  = {white, c},
cell{3}{1} = {r=4,c=1}{c,cellcol},
cell{7}{1} = {r=4,c=1}{c,cellcol},
cell{11}{1} = {r=4,c=1}{c,cellcol},
cell{14}{10} = {first}, cell{14}{9} = {second}, cell{14}{5} = {second},
cell{13}{10} = {first}, cell{13}{9} = {second}, cell{13}{5} = {first},
cell{12}{10} = {second}, cell{12}{5} = {first},
cell{11}{10} = {first}, cell{11}{5} = {second},
cell{10}{10} = {first}, cell{10}{9} = {second},
cell{9}{10} = {first}, cell{9}{5} = {second},
cell{8}{10} = {first}, cell{8}{5} = {second},
cell{7}{10} = {first}, cell{7}{5} = {second},
cell{6}{10} = {first}, cell{6}{9} = {second},
cell{5}{10} = {first}, cell{5}{9} = {second},
cell{4}{10} = {first}, cell{4}{9} = {second},
cell{3}{10} = {first}, cell{3}{9} = {second},
colsep=3pt, rowsep=0.5pt,hspan=minimal
                }
{Metric} & &Vanilla & \small{Temp Scaling} & {DEns}& {MCD} & \small{LL-Dropout}& {SVI} & {SVI-AvUC} & {Ours}
\\
&&&&&&&&&
\\ \hline
ECE $\downarrow$ & lower quartile & 0.124 & 0.096 & 0.050 & 0.078 & 0.093 & 0.072 & 0.032 & 0.022
\\
   & median & 0.174 & 0.139 & 0.090 & 0.134 & 0.145 & 0.114 & 0.045 & 0.038\\
   & mean & 0.194 & 0.160 & 0.088 & 0.153 & 0.161 & 0.119 & 0.054 & 0.044\\
& upper quartile & 0.274 & 0.236 & 0.126 & 0.219 & 0.236 & 0.172 & 0.070 & 0.063\\ \hline
 NLL $\downarrow$   & lower quartile & 4.635 & 4.53 & 4.035 & 4.699 & 4.563 & 4.322 & 4.164 & 4.014
 \\
   & median & 5.115 & 4.993 & 4.624 & 5.093 & 5.034 & 4.853 & 4.823 & 4.617
   \\
   & mean & 5.234 & 5.091 & 4.604 & 5.553 & 5.201 & 4.865 & 4.707 & 4.352
   \\
& upper quartile & 6.292 & 6.165 & 5.893 & 6.522 & 6.342 & 6.034 & 5.778 & 4.987
\\ \hline 
 Brier $\downarrow$  & lower quartile & 0.941 & 0.926 & 0.877 & 0.933 & 0.923 & 0.906 & 0.883 & 0.868
 \\
   & median & 0.987 & 0.970 & 0.922 & 0.967 & 0.969 & 0.943 & 0.935 & 0.925
   \\
   & mean & 0.964 & 0.945 & 0.888 & 0.961 & 0.947 & 0.922 & 0.900 & 0.887
   \\
& upper quartile & 1.052 & 1.027 & 0.989 & 1.025 & 1.025 & 1.013 & 0.985 & 0.949\\
\hline
\end{tblr}
}
\vspace{-10pt}
\end{table}

\paragraph{b. Calibration under distribution shifts.}
Following our observation that our uncertainty estimates are effective in rejecting outliers in \ref{tab:Imagenet}, here we study if they can be leveraged to calibrate ImageNet models under distribution shifts. To calibrate a classifier, we simply scale the logits of the mean by the uncertainties as follows: $\boldsymbol\mu_{\text{calib.}} = \boldsymbol\mu (1-\bar{\boldsymbol\sigma})$, where $\bar{\boldsymbol\sigma}$ is the standard deviation estimated from \eqref{eqn:deluq_pred}, where once again for classification we simply compute the average of standard deviation across all the 1000 classes, followed by min-max normalization to $[0,1)$. This simple scaling of the mean reflects our prior belief -- a highly certain prediction must remain unchanged, whereas an uncertain one gets tempered down. Note, this scaling is applied to logits from all classes and hence the accuracy of the mean remains unchanged before or after calibration. We evaluate how calibrated the predictions are using three commonly used metrics -- Calibration error (ECE), negative log likelihood (NLL), and Brier Score. We use the same ResNet-50 classifier trained on ImageNet as before, and measure calibration for predictions on 16 different ImageNet-C corruptions at severity $5$. We list the corruptions and show examples of them in the appendix. We report the $25^\text{th}$ (lower quartile), $50^\text{th}$ (median) and $75^\text{th}$ (upper quartile) along with the mean of the three metrics across $16$ corruptions in Table \ref{tab:Imagenet_calibration}. Across all measures we see that \name is able to calibrate models better, even in comparison to state-of-the-art approaches that use an explicit calibration objective to adjust the prediction probabilities, further validating the quality of its uncertainty estimates. 

% When $f$ is Lipschitz continuous, i.e., $\|f(\mathrm{x}) - f(\mathrm{x}^{\prime})\| \leq c \|\mathrm{x} - \mathrm{x}^{\prime}\|$, and its first- and second-order information are accessible, this can be solved using first-order optimization methods such as stochastic gradient descent (SGD)~\cite{bottou2012stochastic} or second-order methods such as L-BFGS~\cite{liu1989limited}. However, in practice, while $f$ can be explicitly evaluated for any $\mathrm{x}$, its first- and second-order information are unknown, thus making such an optimization very challenging. Furthermore,
% At the core of AI-powered applications in science and engineering lies the need to perform design optimization for maximizing a chosen target objective, and to enable automated exploration in high-dimensional parameter spaces.

\paragraph{c. Sequential Optimization.} Denoting a high-dimensional function as $\mathrm{f}: \mathcal{D} \rightarrow \mathbb{R}$, our goal is to solve the following optimization problem: $\mathrm{x}^* = arg \max_{\mathrm{x} \in \mathcal{D}} \mathrm{f}(\mathrm{x})$. Here, $\mathcal{D}$ refers to a \textit{bounded} design space comprising $D$ different parameters with their corresponding value ranges $[\ell_d, h_d], \forall d = 1 \cdots D$. The high computational or financial cost of evaluating $\mathrm{f}$ (invoking a simulator or running an experiment) motivates the additional objective of minimizing the number of evaluations. 

% Commonly referred to as \textit{black-box optimization}~\cite{audet2017derivative}, this formulation is adopted in applications ranging from drug design~\cite{schneider2020rethinking} to additive manufacturing~\cite{wang2020machine} and optimizing financial investments~\cite{gonzalvez2019financial} to hyper-parameter tuning in neural networks~\cite{ren2021comprehensive}. 
% \begin{equation}
%     \mathrm{x}^* = arg \max_{\mathrm{x} \in \mathcal{D}} \mathrm{f}(\mathrm{x}).
%     \label{eq:basic}
% \end{equation}

Given the high-dimensional nature of the design spaces, a simple brute-force search or even space-filling random sample designs~\cite{kailkhura2018spectral} often require significantly large sample sizes to identify the optima, thus motivating the use of \textit{sequential optimization} strategies. In particular, Bayesian Optimization (BO) techniques based on statistical surrogates (e.g., Gaussian processes) form an important class of solutions~\cite{shahriari2015taking}. In a nutshell, given an initial experiment design and their function evaluations, $\{\mathrm{x}_i, \mathrm{f}(\mathrm{x}_i)\}_{i=0}^{n_0}$, sequential optimization techniques incrementally select candidates to achieve the so-called \textit{exploration-exploitation} trade-off using an appropriate \textit{acquisition} function~\cite{snoek2012practical}. In this study, we use the popular expected improvement (EI) score to perform candidate selection.

\begin{table}[tb!]
\renewcommand{\arraystretch}{1.25}
\centering
\caption{\small{\textbf{Sequential optimization:} We rigorously evaluate the performance of different uncertainty estimators on a suite of black-box functions and report the AUC metric ($\uparrow$) averaged across multiple random seeds and trials. In each case, we also indicate the number of initial samples and optimization steps.}}
\resizebox{1.\textwidth}{!}{
\begin{tabular}{|c||c||c||c||c||c||c||c||c|}
\hhline{|-||-||-||-||-||-||-||-||-|}
\rowcolor[HTML]{404040} 
\textcolor{white}{\textbf{Function}} & \textcolor{white}{\textbf{Dim.}} & \textcolor{white}{\textbf{Init.}} & \textcolor{white}{\textbf{Steps}} & \textcolor{white}{\textbf{GP}} & \textcolor{white}{\textbf{MCD}} & \textcolor{white}{\textbf{BNN}} & \textcolor{white}{\textbf{DEns}} & \textcolor{white}{\textbf{Ours}}                 \\ 
\hhline{=::=::=::=::=::=::=::=::=:}
Multi Optima & $1$ & $5$ & $25$ & $0.51 \pm 0.2$ & $0.45 \pm 0.16$ & \cellcolor[HTML]{FFDBB0}$0.64 \pm 0.12$ & $0.28 \pm 0.17$ & \cellcolor[HTML]{C6F7C6}$0.73 \pm 0.09$\\ \hhline{=::=::=::=::=::=::=::=::=:}
Ackley & $2$ & $5$ & $25$& $0.23 \pm 0.08$ & \cellcolor[HTML]{FFDBB0}$0.76 \pm 0.03$ & $0.71 \pm 0.1$ & $0.75 \pm 0.04$ & \cellcolor[HTML]{C6F7C6}$0.83 \pm 0.03$\\ \hhline{=::=::=::=::=::=::=::=::=:}
Beale & $2$ & $5$ & $25$& $0.64 \pm 0.31$ & $0.55 \pm 0.22$ & $0.27 \pm 0.17$ & \cellcolor[HTML]{FFDBB0}$0.81 \pm 0.03$ & \cellcolor[HTML]{C6F7C6}$0.85 \pm 0.04$\\ \hhline{=::=::=::=::=::=::=::=::=:}
Booth & $2$ & $5$ & $25$& $0.39 \pm 0.21$ & $0.55 \pm 0.14$ & $0.3 \pm 0.2$ & \cellcolor[HTML]{FFDBB0}$0.68 \pm 0.06$ & \cellcolor[HTML]{C6F7C6}$0.79 \pm 0.04$\\ \hhline{=::=::=::=::=::=::=::=::=:}
Branin & $2$ & $5$ & $25$& $0.35 \pm 0.28$ & $0.28 \pm 0.19$ & $0.22 \pm 0.14$ & \cellcolor[HTML]{FFDBB0}$0.46 \pm 0.1$ & \cellcolor[HTML]{C6F7C6}$0.67 \pm 0.06$\\ \hhline{=::=::=::=::=::=::=::=::=:}
Bukin & $2$ & $5$ & $25$& $0.36 \pm 0.12$ & $0.55 \pm 0.07$ & $0.38 \pm 0.11$ & \cellcolor[HTML]{FFDBB0}$0.59 \pm 0.11$ & \cellcolor[HTML]{C6F7C6}$0.76 \pm 0.1$\\ \hhline{=::=::=::=::=::=::=::=::=:}
Camel & $2$ & $5$ & $25$& $0.83 \pm 0.08$ & \cellcolor[HTML]{FFDBB0}$0.86 \pm 0.06$ & $0.84 \pm 0.03$ & $0.83 \pm 0.07$ & \cellcolor[HTML]{C6F7C6}$0.89 \pm 0.03$\\ \hhline{=::=::=::=::=::=::=::=::=:}
Dropwave & $2$ & $5$ & $25$& \cellcolor[HTML]{FFDBB0}$0.68 \pm 0.15$ & $0.57 \pm 0.18$ & $0.67 \pm 0.13$ & $0.67 \pm 0.11$ & \cellcolor[HTML]{C6F7C6}$0.79 \pm 0.14$\\ \hhline{=::=::=::=::=::=::=::=::=:}
Griewank & $2$ & $5$ & $25$& \cellcolor[HTML]{FFDBB0}$0.83 \pm 0.02$ & $0.74 \pm 0.04$ & $0.59 \pm 0.17$ & $0.7 \pm 0.14$ & \cellcolor[HTML]{C6F7C6}$0.86 \pm 0.03$\\ \hhline{=::=::=::=::=::=::=::=::=:}
Holder & $2$ & $5$ & $25$& $0.12 \pm 0.06$ & $0.36 \pm 0.28$ & $0.36 \pm 0.37$ & \cellcolor[HTML]{FFDBB0}$0.39 \pm 0.29$ & \cellcolor[HTML]{C6F7C6}$0.57 \pm 0.07$\\ \hhline{=::=::=::=::=::=::=::=::=:}
Levi N.13 & $2$ & $5$ & $25$& $0.26 \pm 0.26$ & \cellcolor[HTML]{FFDBB0}$0.75 \pm 0.1$ & $0.7 \pm 0.1$ & $0.6 \pm 0.11$ & \cellcolor[HTML]{C6F7C6}$0.87 \pm 0.07$\\ \hhline{=::=::=::=::=::=::=::=::=:}
Levy & $2$ & $5$ & $25$& $0.57 \pm 0.18$ & \cellcolor[HTML]{FFDBB0}$0.61 \pm 0.32$ & $0.55 \pm 0.16$ & $0.59 \pm 0.16$ & \cellcolor[HTML]{C6F7C6}$0.83 \pm 0.03$\\ \hhline{=::=::=::=::=::=::=::=::=:}
Hartmann & $3$ & $5$ & $25$& \cellcolor[HTML]{FFDBB0}$0.57 \pm 0.07$ & $0.49 \pm 0.11$ & $0.46 \pm 0.16$ & $0.53 \pm 0.15$ & \cellcolor[HTML]{C6F7C6}$0.68 \pm 0.07$\\ \hhline{=::=::=::=::=::=::=::=::=:}
Ackley & $4$ & $10$ & $25$& \cellcolor[HTML]{FFDBB0}$0.17 \pm 0.02$ & $0.06 \pm 0.04$ & $0.1 \pm 0.03$ & $0.14 \pm 0.02$ & \cellcolor[HTML]{C6F7C6}$0.59 \pm 0.05$\\ \hhline{=::=::=::=::=::=::=::=::=:}
Griewank & $4$ & $10$ & $25$& $0.37 \pm 0.08$ & \cellcolor[HTML]{FFDBB0}$0.47 \pm 0.06$ & $0.39 \pm 0.05$ & $0.43 \pm 0.07$ & \cellcolor[HTML]{C6F7C6}$0.69 \pm 0.07$\\ \hhline{=::=::=::=::=::=::=::=::=:}
Levy & $4$ & $10$ & $25$& $0.1 \pm 0.08$ & \cellcolor[HTML]{FFDBB0}$0.4 \pm 0.3$ & $0.27 \pm 0.21$ & $0.21 \pm 0.1$ & \cellcolor[HTML]{C6F7C6}$0.62 \pm 0.2$\\ \hhline{=::=::=::=::=::=::=::=::=:}
Hartmann & $6$ & $10$ & $25$& $0.15 \pm 0.01$ & \cellcolor[HTML]{FFDBB0}$0.2 \pm 0.08$ & $0.1 \pm 0.05$ & $0.15 \pm 0.04$ & \cellcolor[HTML]{C6F7C6}$0.27 \pm 0.15$\\ \hhline{=::=::=::=::=::=::=::=::=:}
Ackley & $8$ & $10$ & $50$& $0.06 \pm 0.09$ & $0.09 \pm 0.13$ & $0.08 \pm 0.12$ & \cellcolor[HTML]{FFDBB0}$0.11 \pm 0.05$ & \cellcolor[HTML]{C6F7C6}$0.36 \pm 0.09$\\ \hhline{=::=::=::=::=::=::=::=::=:}
Griewank & $8$ & $10$ & $50$& $0.07 \pm 0.02$ & $0.12 \pm 0.13$ & $0.08 \pm 0.07$ & \cellcolor[HTML]{FFDBB0}$0.19 \pm 0.07$ & \cellcolor[HTML]{C6F7C6}$0.32 \pm 0.11$\\ \hhline{=::=::=::=::=::=::=::=::=:}
Levy & $8$ & $10$ & $50$& $0.12 \pm 0.04$ & \cellcolor[HTML]{FFDBB0}$0.16 \pm 0.08$ & $0.11 \pm 0.07$ & $0.13 \pm 0.07$ & \cellcolor[HTML]{C6F7C6}$0.47 \pm 0.03$\\ 
\hhline{=::=::=::=::=::=::=::=::=:}
\cellcolor[HTML]{BABABA} Avg. Rank &\cellcolor[HTML]{BABABA}- & \cellcolor[HTML]{BABABA}-& \cellcolor[HTML]{BABABA}-& \cellcolor[HTML]{BABABA}$3.8$ & \cellcolor[HTML]{BABABA} $3.0$ & \cellcolor[HTML]{BABABA} $4.1$ & \cellcolor[HTML]{BABABA} $2.95$ & \cellcolor[HTML]{BABABA} $1.0$\\ 
%\hhline{=::=::=::=::=::=::=::=::=:}
%ICF Hydra & $8$ & $10$ & $50$& $0.49 \pm 0.07$ & $0.69 \pm 0.08$ & $0.47 \pm 0.23$ & \cellcolor[HTML]{FFDBB0}$0.7 %\pm 0.09$ & \cellcolor[HTML]{C6F7C6}$0.78 \pm 0.15$\\ 
\hhline{|-||-||-||-||-||-||-||-||-|}
\end{tabular}
}
\label{tab:seqopt}
\vspace{0pt}
\end{table}

% In each step of this optimization, we approximate the function $f$ using the samples observed so far to obtain the surrogate $\hat{f}$. Assuming that our goal is to maximize $f$, one can acquire more samples in regimes of $\mathcal{D}$ where the mean estimate from the surrogate is high (\textit{exploitation}) or the uncertainty is large (\textit{exploration}). In order to balance between these two objectives and to guide the progressive search for the optima, BO utilizes an appropriate \textit{acquisition} function~\cite{snoek2012practical}. In this study, we use the popular expected improvement (EI) score to perform candidate selection.
% % \begin{align}
% \nonumber
% \text{aq}_{\text{EI}} & \coloneqq 
% \begin{cases}
%   (\mu(\mathrm{x}) - f(\mathrm{x}^+) - \xi) \Phi(\mathrm{Z}) + \sigma(\mathrm{x}) \phi(\mathrm{Z}) & \text{if $\sigma(\mathrm{x})>0$} \\
%   0 & \text{if $\sigma(\mathrm{x})=0$}
% \end{cases} \\
% &\text{where } \mathrm{Z} = \frac{(\mu(\mathrm{x}) - f(\mathrm{x}^+) - \xi)}{\sigma(\mathrm{x})}.
%     \label{eq:ei}
% \end{align}Here, $\mu(\mathrm{x})$ and $\sigma(\mathrm{x})$ are the mean and uncertainty estimates from the surrogate $\hat{f}$ for any sample, and $f(\mathrm{x}^+)$ is the best known function value so far during any iteration of the optimization. Further, $\Phi(.)$ and $\phi(.)$ denote the cumulative distribution and probability density functions corresponding to the normal distribution. Finally, the hyper-parameter $\xi$ controls the exploration-exploitation trade-off.

\begin{wrapfigure}{r}{5cm}
\vspace{-0.1in}
\centering
\includegraphics[width=0.35\textwidth]{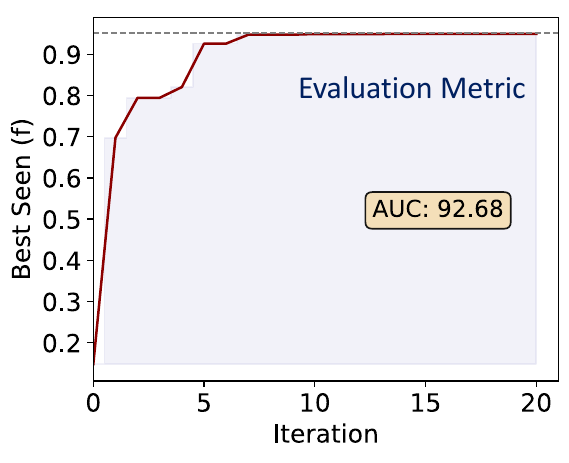}
\caption{\small{\textbf{Area under the curve (AUC) metric} for evaluating sequential optimization performance.}}
\vspace{-0.1in}
\label{fig:auc}
\end{wrapfigure}\noindent \underline{\textit{Setup}}: In this experiment, we consider a large suite of black-box optimization functions with varying dimensionality ($1$ to $8$) and complexity. We provide definitions and details of these functions used in our study in the appendix. We also perform an experiment with a pre-trained generative model (GAN) trained on MNIST handwritten digits, wherein we perform optimization in the $100-$D latent space $\mathcal{Z}$ such that thickness of the resulting digit is maximized: $\sum_i \mathbb{I}(x_i > 0), \forall i$, where $\mathbb{I}$ denotes the identity function. We use the following baseline uncertainty estimation approaches in our study: (i) Gaussian processes (GP); (ii) Monte-Carlo dropout (MCD); (iii) Bayesian neural networks (BNN) trained via variational inferencing; and (iv) deep ensembles (DEns). For all neural network surrogates, we computed positional embeddings (sinusoidal) of the raw parameter inputs prior to building a fully-connected network with $4$ hidden layers each containing $128$ neurons and ReLU activation. All methods were trained with the same set of hyperparameters: Adam optimizer learning rate $1e-4$ and $500$ epochs, except for BNN, which required $1000$ epochs for convergence. With MCD, we used $50$ forward passes at test time for each sample to obtain the uncertainties. Finally, with $\Delta-$UQ, we set the number of anchors for inferencing as $\min(20, n)$, where $n$ is the number of samples in the observed dataset in any iteration.

\begin{figure}[htb!]
    \centering
    \includegraphics[width = 0.99\linewidth]{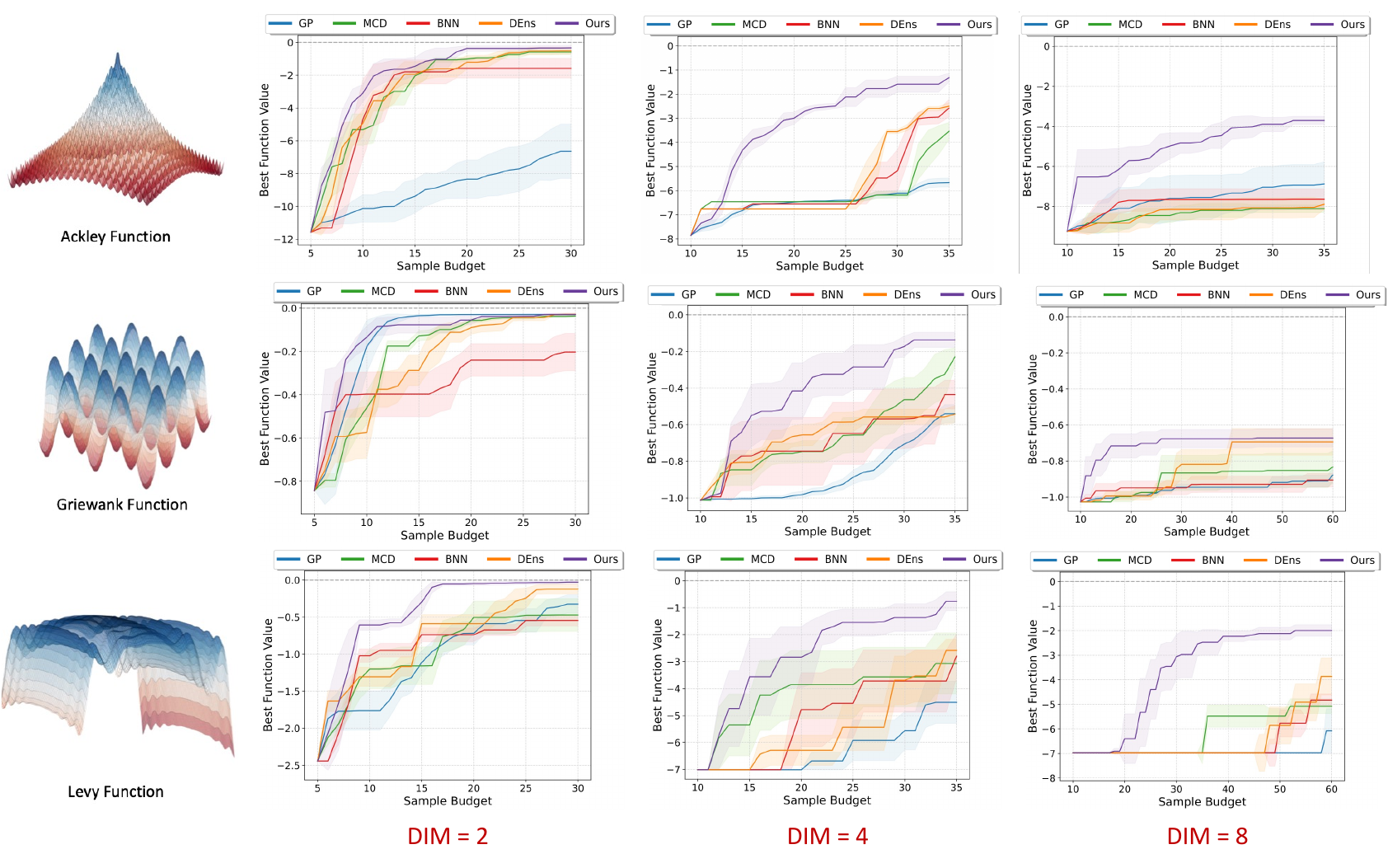}
    \caption{\small{\textbf{Convergence curves obtained with different uncertainty estimation methods:} We show the best function value achieved for three different functions at dimensions 2, 4 and 8 respectively (for 1 random seed, 5 trials). We find that $\Delta-$UQ consistently outperforms all other baselines. The effectiveness of our approach in producing meaningful uncertainties at small sample sizes becomes more apparent as dimensionality increases.}}
    \label{fig:opt_dim}
    \vspace{-5pt}
\end{figure}
\begin{figure}[htb!]
    \centering
    \includegraphics[width = 0.85\linewidth]{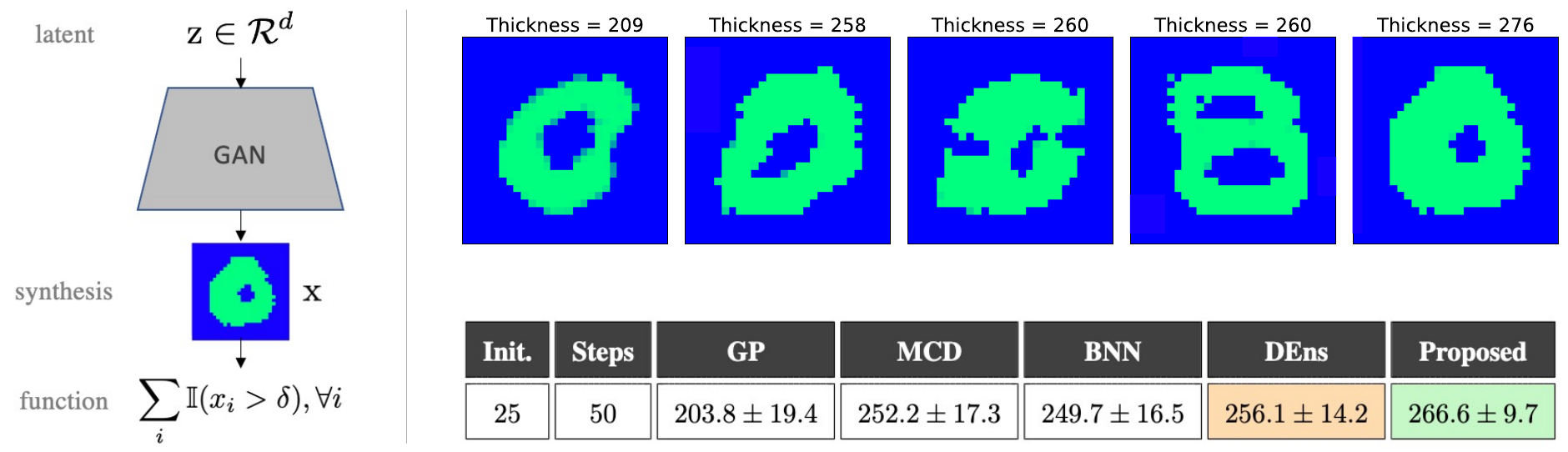}
    \caption{\small{\textbf{GAN-based optimization}: $\Delta-$UQ consistently produces images with higher function values (thickness) for the same sampling budget, when compared to existing baseline methods.}}
    \label{fig:mnist}
    \vspace{-5pt}
\end{figure}

The DEns model was constructed using $5$ constituent members (increasing this did not provide any benefits), each trained with a different initialization. The number of initial samples and the number of steps in the sequential optimization were set to be the same across all methods. In each round of the Bayesian optimization, we used $10,000$ samples for initialization and $15$ restarts (i.e., starting points for multistart acquisition function optimization), and finally one candidate ($q = 1$) was evaluated with the black-box function and added to the observed dataset. We performed experiments with $5$ random seeds (different initializations), each for $5$ independent trials. Since the goal is to reach the global optima with the fewest number of samples, we use the widely adopted area under the \textit{iteration vs best achieved function value} curve to obtain a holistic evaluation of different approaches (see Figure \ref{fig:auc}).

% \begin{table}[t!
\noindent \underline{\textit{Results}}: From table \ref{tab:seqopt}, we find that $\Delta-$UQ produces significantly higher AUC scores in comparison to existing baselines, across all benchmark functions. While MCD and DEns behave reasonably well in low dimensions, their performance suffers when we go to higher dimensions (see figure \ref{fig:opt_dim}). Furthermore, we find that the performance of BNN is generally lower due to the inherent small samples sizes that we operate in. Finally, as shown in figure \ref{fig:mnist}, our approach consistently achieves higher function values in the MNIST GAN-based optimization, thus validating the quality of the uncertainties produced via anchoring.

%Even in the case of the real-world ICF Hydra simulator, $\Delta-UQ$ achieves the best convergence characteristics with an average AUC score of $0.78$. 

% \renewcommand{\arraystretch}{1.8}
% \centering
% \caption{MNIST result}
% \resizebox{\textwidth}{!}{
% \begin{tabular}{|c||c||c||c||c||c||c|}
% \hhline{|-||-||-||-||-||-||-|}
% \rowcolor[HTML]{404040} 
% \textcolor{white}{\textbf{Init.}} & \textcolor{white}{\textbf{Steps}} & \textcolor{white}{\textbf{GP}} & \textcolor{white}{\textbf{MCD}} & \textcolor{white}{\textbf{BNN}} & \textcolor{white}{\textbf{DEns}} & \textcolor{white}{\textbf{Proposed}}                 \\ 
% \hhline{=::=::=::=::=::=::=:}
%  $25$ & $50$ & $203.8 \pm 19.4$ & $252.2 \pm 17.3$ & $249.7 \pm 16.5$ & \cellcolor[HTML]{FFDBB0}$256.1 \pm 14.2$ & \cellcolor[HTML]{C6F7C6}$266.6 \pm 9.7$\\
% \hhline{|-||-||-||-||-||-||-|}
% \end{tabular}
% }
% \end{table}

%% file: discussion.tex
\section{Broader Impact}
We presented a simple, scalable, and accurate single model uncertainty estimator that outperforms many existing techniques. Uncertainty quantification (UQ) plays a significant role in a variety of applications in science and engineering, from developing safeguards in critical applications such as healthcare, security, or finance to exploration of novel design spaces using sequential optimization. Our paper addresses both of these applications using demonstrative problems to showcase the potential capabilities. Our goal is to present a fundamental approach for uncertainty estimation that maybe applicable in different kinds of problems and settings. Due to its fundamental nature, we do not foresee misuse directly. However, we note that our method does not mitigate biases that may come with the training data, which could potentially affect the quality of uncertainties themselves.

%% file: checklist.tex
%%% BEGIN INSTRUCTIONS %%%
% The checklist follows the references.  Please
% read the checklist guidelines carefully for information on how to answer these
% questions.  For each question, change the default \answerTODO{} to \answerYes{},
% \answerNo{}, or \answerNA{}.  You are strongly encouraged to include a {\bf
% justification to your answer}, either by referencing the appropriate section of
% your paper or providing a brief inline description.  For example:
% \begin{itemize}
%   \item Did you include the license to the code and datasets? \answerYes{See Section~\ref{gen_inst}.}
%   \item Did you include the license to the code and datasets? \answerNo{The code and the data are proprietary.}
%   \item Did you include the license to the code and datasets? \answerNA{}
% \end{itemize}
% Please do not modify the questions and only use the provided macros for your
% answers.  Note that the Checklist section does not count towards the page
% limit.  In your paper, please delete this instructions block and only keep the
% Checklist section heading above along with the questions/answers below.
% %%% END INSTRUCTIONS %%%
\section*{Checklist}
\begin{enumerate}

\item For all authors...
\begin{enumerate}
  \item Do the main claims made in the abstract and introduction accurately reflect the paper's contributions and scope?
    \answerYes{}
  \item Did you describe the limitations of your work?
    \answerYes{}
  \item Did you discuss any potential negative societal impacts of your work?
    \answerYes{}
  \item Have you read the ethics review guidelines and ensured that your paper conforms to them?
    \answerYes{}
\end{enumerate}

\item If you are including theoretical results...
\begin{enumerate}
  \item Did you state the full set of assumptions of all theoretical results?
    \answerYes{}
        \item Did you include complete proofs of all theoretical results?
    \answerYes{Main claims are presented in the paper, further derivations are given in the supplement}
\end{enumerate}

\item If you ran experiments...
\begin{enumerate}
  \item Did you include the code, data, and instructions needed to reproduce the main experimental results (either in the supplemental material or as a URL)?
    \answerYes{}
  \item Did you specify all the training details (e.g., data splits, hyperparameters, how they were chosen)?
    \answerYes{}
        \item Did you report error bars (e.g., with respect to the random seed after running experiments multiple times)?
    \answerYes{}
        \item Did you include the total amount of compute and the type of resources used (e.g., type of GPUs, internal cluster, or cloud provider)?
    \answerNA{}
\end{enumerate}

\item If you are using existing assets (e.g., code, data, models) or curating/releasing new assets...
\begin{enumerate}
  \item If your work uses existing assets, did you cite the creators?
    \answerYes{}
  \item Did you mention the license of the assets?
    \answerNA{We are using standard open sourced datasets and models}
  \item Did you include any new assets either in the supplemental material or as a URL?
    \answerNo{}
  \item Did you discuss whether and how consent was obtained from people whose data you're using/curating?
    \answerNA{}
  \item Did you discuss whether the data you are using/curating contains personally identifiable information or offensive content?
    \answerNA{}
\end{enumerate}

\item If you used crowdsourcing or conducted research with human subjects...
\begin{enumerate}
  \item Did you include the full text of instructions given to participants and screenshots, if applicable?
    \answerNA{}
  \item Did you describe any potential participant risks, with links to Institutional Review Board (IRB) approvals, if applicable?
    \answerNA{}
  \item Did you include the estimated hourly wage paid to participants and the total amount spent on participant compensation?
    \answerNA{}
\end{enumerate}

\end{enumerate}

%%%%%%%%%%%%%%%%%%%%%%%%%%%%%%%%%%%%%%%%%%%%%%%%%%%%%%%%%%%

%% file: supp_anchor_corruption.tex
\section{Derivation for shifted training on NTK}
We continue the derivation from the main here in more detail. Recall, the prediction on a test sample $\mathrm{x}_t$ in the limit as the inner layer widths grow to infinity. It has been shown that (c.f. \cite{lee2019wide,bietti2019inductive}): 
\begin{equation}
\label{suppeqn:ntk_limit}
\centering
f_\infty(\mathrm{x}_t) = f_0(\mathrm{x}_t) - \mathbf{K}_{\mathrm{x}_t\mathbf{X}}\mathbf{K}_{\mathbf{X} \mathbf{X}}^{-1}(f_0(\mathbf{X})-\mathbf{Y}),
\end{equation}where $\mathbf{X}$ is the matrix of all training data samples. As before, we consider the case where the domain is shifted by $\mathrm{c}$. Using \eqref{suppeqn:ntk_limit}:
\begin{align}
    &f_\infty(\mathrm{x}_t-\mathrm{c}) = f_0(\mathrm{x}_t-\mathrm{c}) - \mathbf{K}_{(\mathrm{x}_t-\mathrm{c})(\mathbf{X}-\mathrm{c})}\mathbf{K}_{(\mathbf{X}-\mathrm{c}) (\mathbf{X}-\mathrm{c})}^{-1}(f_0(\mathbf{X}-\mathrm{c})-\mathbf{Y}) \notag &\\
    &\approx f_0(\mathrm{x}_t-\mathrm{c}) - (\mathbf{K}_{\mathrm{x}_t\mathbf{X}} - \Gamma_{\mathrm{x}_t, \mathbf{X},\mathrm{c}})(\mathbf{K}_{\mathbf{X}\mathbf{X}} - \Gamma_{\mathbf{X}, \mathbf{X},\mathrm{c}})^{-1}(f_0(\mathbf{X}-\mathrm{c})-\mathbf{Y})\label{suppeqn:shifted_ntk_1}
\end{align} 
Where we utilize Woodbury's Identity \cite{woodbury1950inverting} for expanding the inverse of the difference between two matrices as:

\begin{equation}
\label{suppeqn:woodbury}
    (A - B)^{-1} = A^{-1} + \sum_{m=1}^\infty (A^{-1}B)^mA^{-1}
\end{equation}
Using \eqref{suppeqn:woodbury}, we can expand \eqref{suppeqn:shifted_ntk_1} as:

\begin{align}
\label{suppeqn:expanded_woodbury}
     &= f_0(\mathrm{x}_t-\mathrm{c}) - (\mathbf{K}_{\mathrm{x}_t\mathbf{X}} - \Gamma_{\mathrm{x}_t, \mathbf{X},\mathrm{c}})\left(\mathbf{K}_{\mathbf{X}\mathbf{X}}^{-1} + \sum_{m=1}^\infty (\mathbf{K}_{\mathbf{X}\mathbf{X}}^{-1}\Gamma_{\mathbf{X}, \mathbf{X},\mathrm{c}})^m\mathbf{K}_{\mathbf{X}\mathbf{X}}^{-1}\right)(f_0(\mathbf{X}-\mathrm{c})-\mathbf{Y}) &\\
     &= f_0(\mathrm{x}_t-\mathrm{c}) - (\mathbf{K}_{\mathrm{x}_t\mathbf{X}} - \Gamma_{\mathrm{x}_t, \mathbf{X},\mathrm{c}})\mathbf{K}_{\mathbf{X}\mathbf{X}}^{-1}(f_0(\mathbf{X}-\mathrm{c})-\mathbf{Y}) ~~- \tag{contd.} &\\ 
     &(\mathbf{K}_{\mathrm{x}_t\mathbf{X}} - \Gamma_{\mathrm{x}_t, \mathbf{X},\mathrm{c}})\sum_{m=1}^\infty (\mathbf{K}_{\mathbf{X}\mathbf{X}}^{-1}\Gamma_{\mathbf{X}, \mathbf{X},\mathrm{c}})^m\mathbf{K}_{\mathbf{X}\mathbf{X}}^{-1}(f_0(\mathbf{X}-\mathrm{c})-\mathbf{Y}) \notag &\\
     &= \underbrace{f_0(\mathrm{x}_t-\mathrm{c}) - \mathbf{K}_{\mathrm{x}_t\mathbf{X}}\mathbf{K}_{\mathbf{X}\mathbf{X}}^{-1}(f_0(\mathbf{X}-\mathrm{c})-\mathbf{Y})}_{\text{first}} ~~- \tag{contd.}  &\\
     &\Gamma_{\mathrm{x}_t, \mathbf{X},\mathrm{c}}\mathbf{K}_{\mathbf{X}\mathbf{X}}^{-1}(f_0(\mathbf{X}-\mathrm{c})-\mathbf{Y}) - (\mathbf{K}_{\mathrm{x}_t\mathbf{X}} - \Gamma_{\mathrm{x}_t, \mathbf{X},\mathrm{c}})\sum_{m=1}^\infty (\mathbf{K}_{\mathbf{X}\mathbf{X}}^{-1}\Gamma_{\mathbf{X}, \mathbf{X},\mathrm{c}})^m\mathbf{K}_{\mathbf{X}\mathbf{X}}^{-1}(f_0(\mathbf{X}-\mathrm{c})-\mathbf{Y}) \label{suppeqn:expanded_woodbury_1}
\end{align}
Next, we consider expanding the first term in \eqref{suppeqn:expanded_woodbury_1}. Since the only term dependent on $\mathrm{c}$ is the evaluation of the network with the initial weights $\theta_0$, i.e., of the general form $f_0(\mathrm{x}-\mathrm{c})$. We will expand this using a Taylor series approximation by evaluating it at $\mathrm{c}=0$, as following:
\begin{align}
    \label{suppeqn:taylor_expansion}
    f_0(\mathrm{x}-\mathrm{c}) & = f_0(\mathrm{x}) + \mathrm{c}f_0^\prime(\mathrm{x}) + \mathrm{c}^2f_0^{\prime\prime}(\mathrm{x}) + \dots
\end{align}

By substituting \eqref{suppeqn:taylor_expansion} in \eqref{suppeqn:expanded_woodbury_1}, and grouping all the terms that do not depend on $\mathrm{c}$, we can separate the deterministic and stochastic (in $\mathrm{c}$) which gives us our final result as: 
\begin{align}
    &\approx \underbrace{f_0(\mathrm{x}_t) - \mathbf{K}_{\mathrm{x}_t\mathbf{X}}\mathbf{K}_{X\mathbf{X}}^{-1}(f_0(\mathbf{X}) - \mathbf{Y})}_{\text{deterministic for fixed }\boldsymbol{\theta}_0} - \underbrace{g(\mathrm{c}, \mathrm{x}_t, \mathbf{X}, \mathbf{Y})}_{\text{random due to }\mathrm{c}}\label{suppeqn:ntk_anchor_pred}
\end{align}

\begin{figure}[t]
    \centering
    \includegraphics[width=\textwidth]{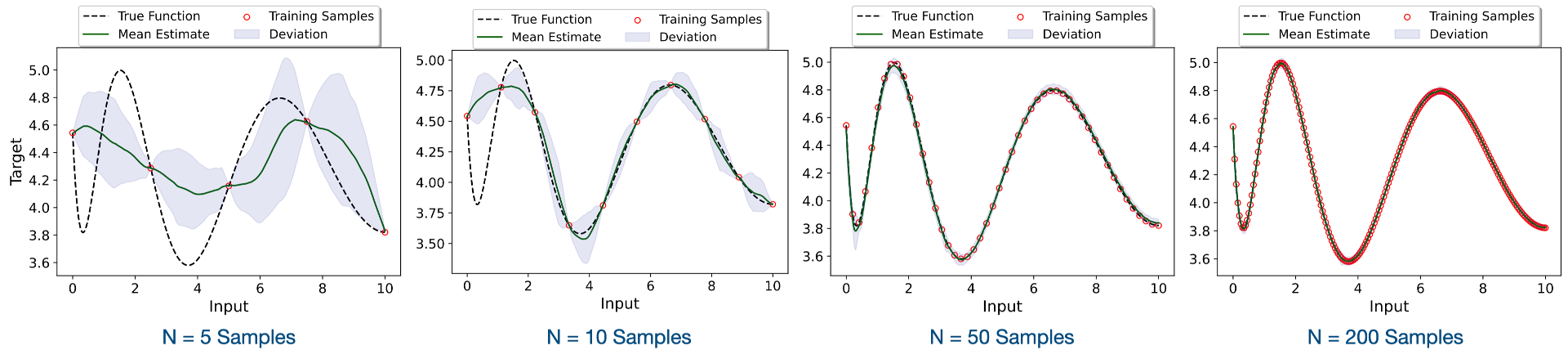}
    \caption{Behavior of the proposed uncertainty estimator as we increase the training sample size for a $1-$D regression example. As expected, as we increase $N$ from $5$ samples to $200$ samples, the prediction uncertainties shrink to trivial estimates, thus emphasizing the ability of our approach in capturing epistemic uncertainties.}
    \label{fig:unc_vs_n}
\end{figure}

\paragraph{Perturbations with different anchors} The analysis above can be easily applied to the case where different anchors are used with different input samples. Let us consider two randomly chosen anchors: $\mathrm{c}_1, \mathrm{c}_2$ and study the dot product between two points shifted using these anchors. 
\begin{align}
\label{eqn:shifted_ntk_diff_anchors}
    [\mathrm{c}_1,\mathrm{x}_1-\mathrm{c}_1]^\top [\mathrm{c}_2,\mathrm{x}_2-\mathrm{c}_2] & = \mathrm{x}_1^\top\mathrm{x}_2 + 2\mathrm{c}_1^\top \mathrm{c}_2 - \mathrm{c}_1^\top \mathrm{x}_2 - \mathrm{c}_2^\top\mathrm{x}_1\\
    &= \mathrm{x}_1^\top\mathrm{x}_2 - \mathrm{c}_1^\top(\mathrm{c}_1\mathrm{c}_2^\top\mathrm{x}_1 + \mathrm{x}_2 - 2\mathrm{c}_2) \notag, 
\end{align}
Where in the last step we exploit the fact that $\mathrm{c}_1,\mathrm{c}_2$ are normalized to be on the hypersphere. We can see that by setting $v = \mathrm{c}_1\mathrm{c}_2^\top\mathrm{x}_1 + \mathrm{x}_2 - 2\mathrm{c}_2$, we can get a similar form of perturbation as $\mathrm{x}_1^\top\mathrm{x}_2 - \mathrm{c}_1^\top v$ by combining all $\mathrm{c}_2$ related terms as before (and equivalently for $\mathrm{c}_1$).

In this paper, we argue that the proposed stochastic data centering technique is effective at estimating epistemic uncertainties with deep networks. For demonstration, let us consider the $1$D regression example showed in Figure \ref{fig:unc_vs_n} and train $\Delta-$UQ models under different train sample sizes ($5, 10, 50$ and $200$ respectively). The figure illustrates the predicted function and the associated uncertainty estimates (shaded region around the predictions). As the training sample size increases, we notice that the uncertainties shrink to trivial values (very close to 0), thus validating that our estimates are strongly correlated with epistemic uncertainties.

\section{Corruptions in the anchoring process}
% It is easy to see that
% \begin{align}
% \label{eq:deluq_dot}
% \left[\mathrm{c},\mathrm{x}_i-\mathrm{c}\right]^\top\left[\mathrm{c},\mathrm{x}_j-\mathrm{c}\right] & = \mathrm{x}_i^\top \mathrm{x}_j - \mathrm{c}^\top(\mathrm{x}_i + \mathrm{x}_j - 2\mathrm{c}) &\\
% \left[\tilde{\mathrm{c}},\mathrm{x}_i-\mathrm{c}\right]^\top\left[\tilde{\mathrm{c}},\mathrm{x}_j-\mathrm{c}\right] & =  \tilde{\mathrm{c}}^\top \tilde{\mathrm{c}} + (\mathrm{x}_i-c)^\top (\mathrm{x}_j-c) &\\
% &=  \tilde{\mathrm{c}}^\top \tilde{\mathrm{c}} + \mathrm{x}_i^\top \mathrm{x}_j - \mathrm{c}^\top \mathrm{x}_i - \mathrm{c}^\top \mathrm{x}_j + \mathrm{c}^\top \mathrm{c}
% \end{align}
When scaling \name to image data, especially using powerful base networks such as ResNets, we observe a consistency training significantly improves the uncertainty estimates. To achieve this, we make the input transformation less trivial as $\mathrm{x} \rightarrow {\mathcal{T}(\mathrm{c}), \mathrm{x}-\mathrm{c}}$, where $\mathcal{T}$ is a standard augmentation technique already used in training such as random crops or blurs. Such that, during inference we set $\mathcal{T} = \mathcal{I}$, to be identity. We apply this corruption once every 10 iterations -- this is a hyper parameter, doing this more frequently makes the process much harder resulting in a worse mean estimate whereas making it less frequently results in uncertainties that are slightly worse. We outline the exact set of corruptions used in the pytorch pseudo code listed below.

\subsection{Pytorch implementation}
Algorithm  \ref{alg:anchoring} lists the Pytorch pseduo-code for training a $\Delta-$UQ for image classification, and a sample inference script in algorithm \ref{alg:inference_pytorch}. The example assumes a model as defined in algorithm \ref{alg:anchoring} and a set of anchors drawn from the training distribution at random. Once predictions per anchor are obtained, the mean and standard deviation are returned as the final prediction of the model, and the corresponding uncertainty on the test samples.  

\begin{algorithm}[htb!]
\centering
  \caption{PyTorch-style example for $\Delta-$UQ with ResNet-50.}
  \label{alg:anchoring}
   
    \definecolor{codeblue}{rgb}{0.25,0.5,0.5}
    \lstset{
      basicstyle=\fontsize{7.2pt}{7.2pt}\ttfamily\bfseries,
      commentstyle=\fontsize{7.2pt}{7.2pt}\color{codeblue},
      keywordstyle=\fontsize{7.2pt}{7.2pt},
    }
\begin{lstlisting}[language=python]
def create_anchored_model(model):
    model.conv1 = nn.Conv2d(in_channels=6, 64)
    return model
    
Tx = transforms.Compose([
    transforms.RandomResizedCrop(size=224),
    transforms.RandomHorizontalFlip(),
    transforms.RandomApply([color_jitter,blurr], p=0.8),
    ])    
## load model and change the first conv layer

model_basic = ResNet50(pre_trained=False,n_class=1000)
model = create_anchored_model(model_basic)

## load datasets, setup optimizer, define criterion etc.
for i, (images, targets) in enumerate(train_loader):
   
    anchors = Shuffle(images)
   
    diff = images-anchors
    
    if i % 10 ==0:
        tx_anchors = Tx(anchors)
    else:
        tx_anchors = anchors

    batch = torch.cat([tx_anchors,diff],axis=1)
    output = model(batch)
    
    loss = criterion(output, target)
    
    optimizer.zero_grad()
    
    loss.backward()
    
    optimizer.step()
\end{lstlisting}
\end{algorithm}

\begin{algorithm}[htb!]
\centering
  \caption{Inference with \name for a classification model}
  \label{alg:inference_pytorch}
   
    \definecolor{codeblue}{rgb}{0.25,0.5,0.5}
    \lstset{
      basicstyle=\fontsize{7.2pt}{7.2pt}\ttfamily\bfseries,
      commentstyle=\fontsize{7.2pt}{7.2pt}\color{codeblue},
      keywordstyle=\fontsize{7.2pt}{7.2pt},
    }
\begin{lstlisting}[language=python]
'''
model       : network trained with anchoring
anchors     : set of randomly chosen anchors (ideally from train dist.)
test_inputs : samples on which predictions are needed
'''
preds = []
for A in anchors:
    D = test_inputs-A 
    X_test = torch.cat([A, D],axis=1)
    y_test = model(X_test)
    preds.append(y_test)
P = torch.cat(preds,0)
mu = P.mean(0)
unc = P.std(0).sum(1) ## sum unc. along classes

\end{lstlisting}
\end{algorithm}

\subsection{ImageNet-C corruptions for OOD and Calibration}
Table \ref{tab:imagenetc} lists the set of corruptions used to construct the ImageNet-C benchmark.
\begin{table}[h!]
\centering
\caption{ImageNet-C corruptions used for the calibration study}
\begin{tabular}{cccc}
\toprule
     brightness & contrast & defocus\_blur & elastic\_transform \\
     fog & frost& gaussian\_blur & gaussian\_noise\\ 
     glass\_blur & glass\_blur& glass\_blur & gaussian\_noise\\ 
     shot\_noise & spatter& speckle\_noise & zoom\_blur\\
\bottomrule
\end{tabular}
\label{tab:imagenetc}
\end{table}
\input{supp_regression}
\input{supp_accuracies}

%% file: supp_regression.tex
\begin{table}[htb!]
\renewcommand{\arraystretch}{1.4}
\centering
\caption{Regression performance evaluation using UCI benchmarks. For each case, we show the negative log-likelihood for the test data obtained using each of the methods. Note, all metrics were computed as an average from $20$ random trials of $0.8-0.2$ train-test split. We followed the experimental setup described in \cite{sun2017learning} and the results for the baselines were obtained from the uncertainty baselines \href{https://github.com/google/uncertainty-baselines/tree/main/baselines/uci}{github page} \cite{nado2021uncertainty}}
\resizebox{0.75\textwidth}{!}{
\begin{tabular}{|c||c||c||c||c||c|}
\hhline{|-||-||-||-||-||-|}
\rowcolor[HTML]{404040} 
\textcolor{white}{\textbf{Function}} & \textcolor{white}{\textbf{MCD}} & \textcolor{white}{\textbf{DEns}} & \textcolor{white}{\textbf{BNN}} & \textcolor{white}{\textbf{PBP}} & \textcolor{white}{\textbf{Proposed}}                 \\ 
\hhline{=::=::=::=::=::=:}
Boston Housing & \cellcolor[HTML]{C6F7C6}2.4 & 6.11 & 3.12 & \cellcolor[HTML]{FFDBB0}2.54 & 2.58\\ 
\hhline{=::=::=::=::=::=:}
Concrete Strength & \cellcolor[HTML]{C6F7C6}2.93 & 3.2 & 3.22 & \cellcolor[HTML]{FFDBB0}3.04 & 3.09\\ 
\hhline{=::=::=::=::=::=:}
Energy Efficiency & 1.21 & \cellcolor[HTML]{FFDBB0}0.61 & 0.93 & 1.01 & \cellcolor[HTML]{C6F7C6}0.56 \\ 
\hhline{=::=::=::=::=::=:}
Kin8nm & -1.14 & -1.17 & -1.03 & \cellcolor[HTML]{C6F7C6}-1.28 & \cellcolor[HTML]{FFDBB0}-1.19 \\ 
\hhline{=::=::=::=::=::=:}
Naval Propulsion & -4.45 & -5.17 & \cellcolor[HTML]{C6F7C6}-6.12 & -4.85 & \cellcolor[HTML]{FFDBB0}-5.86 \\ 
\hhline{=::=::=::=::=::=:}
Power Plant & \cellcolor[HTML]{FFDBB0}2.8 & 3.18 & 2.85 & \cellcolor[HTML]{C6F7C6}2.78 & 2.83 \\ 
\hhline{=::=::=::=::=::=:}
Wine & \cellcolor[HTML]{FFDBB0}0.93 & 0.97 & 1.0 & 0.97 & \cellcolor[HTML]{C6F7C6}0.91 \\ 
\hhline{=::=::=::=::=::=:}
Protein & 2.87 & 3.12 & 2.93 & \cellcolor[HTML]{C6F7C6}2.77 & \cellcolor[HTML]{FFDBB0}2.79 \\ 
\hhline{=::=::=::=::=::=:}
Yacht & 1.25 & \cellcolor[HTML]{FFDBB0}0.73 & 2.01 & 1.64 & \cellcolor[HTML]{C6F7C6}0.66 \\ 
\hhline{=::=::=::=::=::=:}
\rowcolor[HTML]{BABABA} 
Avg. Rank & 2.89 & 3.56 & 4.0 & 2.44 & 2.0 \\ 
\hhline{|-||-||-||-||-||-|}
\end{tabular}
\label{tab:reg}
}
\end{table}

\section{Additional Results: Prediction Performance on UCI Benchmarks}
While our outlier rejection, calibration and sequential optimization experiments clearly established the effectiveness of the proposed uncertainty estimator, we also evaluate the quality of $\Delta-$UQ models, in terms of standard prediction fidelity metrics (regression in this section and classification in next). For this study, we used a suite of regression datasets typically adopted for evaluating deep models, evaluated using the standard experiment protocol in the benchmark defined by \cite{nado2021uncertainty}. For each of the datasets, we fit networks with a single hidden layer ($50$ neurons) and ReLU activation. We trained $20$ independent models with different random $80-20$ train-test splits and report the average performance across the trials. For evaluation, we used the negative log-likelihood metric (lower the better). In addition to our approach, we include the results for MCD, DEns, BNN (variational inferencing) and Probabilistic Backpropagation~\cite{sun2017learning} (with a Matrix-Variate Gaussian prior). Furthermore, for a holistic evaluation, we also report the average rank (across the $5$ methods) from the suite of datasets considered. As showed in Table \ref{tab:reg}, $\Delta-$UQ performs competitively over other baselines, and achieves an average rank of $2.0$. Overall, we find that, in addition to producing high-quality uncertainty estimates, the proposed approach also produces high-quality predictive models.

%% file: supp_accuracies.tex
\section{Additional Results: Prediction Performance on Imagenet and CIFAR-10}
\textbf{ImageNet-C accuracy.} We provide results for classification accuracy of our ImageNet model on the validation set and the distribution shifted variants of ImageNet-C, in table \ref{tab:acc-imagenet-c}. Here, at each severity level (``1''--``5'') we compute the accuracy of the model across all 16 corruptions for that severity outlined in \ref{tab:imagenetc} and report the mean accuracy. We also report the accuracy numbers for the corresponding uncertainty baselines, and see that $\Delta-$UQ does not compromise on accuracy on the clean data, while being highly competitive to Deep Ensembles even on the most severe corruptions. 

\begin{table}[!htb]
\centering
\caption{Accuracy of ResNet-50 Model on ImageNet validation and its distribution shifted variants. }
\label{tab:acc-imagenet-c}
\begin{tabular}{cccccccc}
\toprule
\textbf{Method} & \multicolumn{7}{c}{\textbf{ImageNet-C Dist. Shift Variants (ResNet-50)}}    \\
\cline{2-8}
                & \textbf{val} & \textbf{1} & \textbf{2} & \textbf{3} & \textbf{4} & \textbf{5} & \textbf{Avg.}  \\
\toprule
Vanilla         & 76.1         & 62.5       & 52         & 42         & 30         & 19.5       & 47.8           \\
DEns            & 78.1         & 66         & 56         & 47         & 36         & 22         & 50.05 \\
MC Dropout      & 75           & 60         & 50         & 38         & 29         & 17         & 46             \\
SVI             & 76.1         & 63         & 53         & 43         & 31         & 20         & 48.05          \\
\midrule
$\Delta-$UQ     & 76.1      & 61.7      & 53.1          & 44.2          & 33.2      & 21.8      & 48.95 \\
\bottomrule
\end{tabular}
\end{table}

\begin{figure}[!htb]
    \centering
    \includegraphics[width=\textwidth]{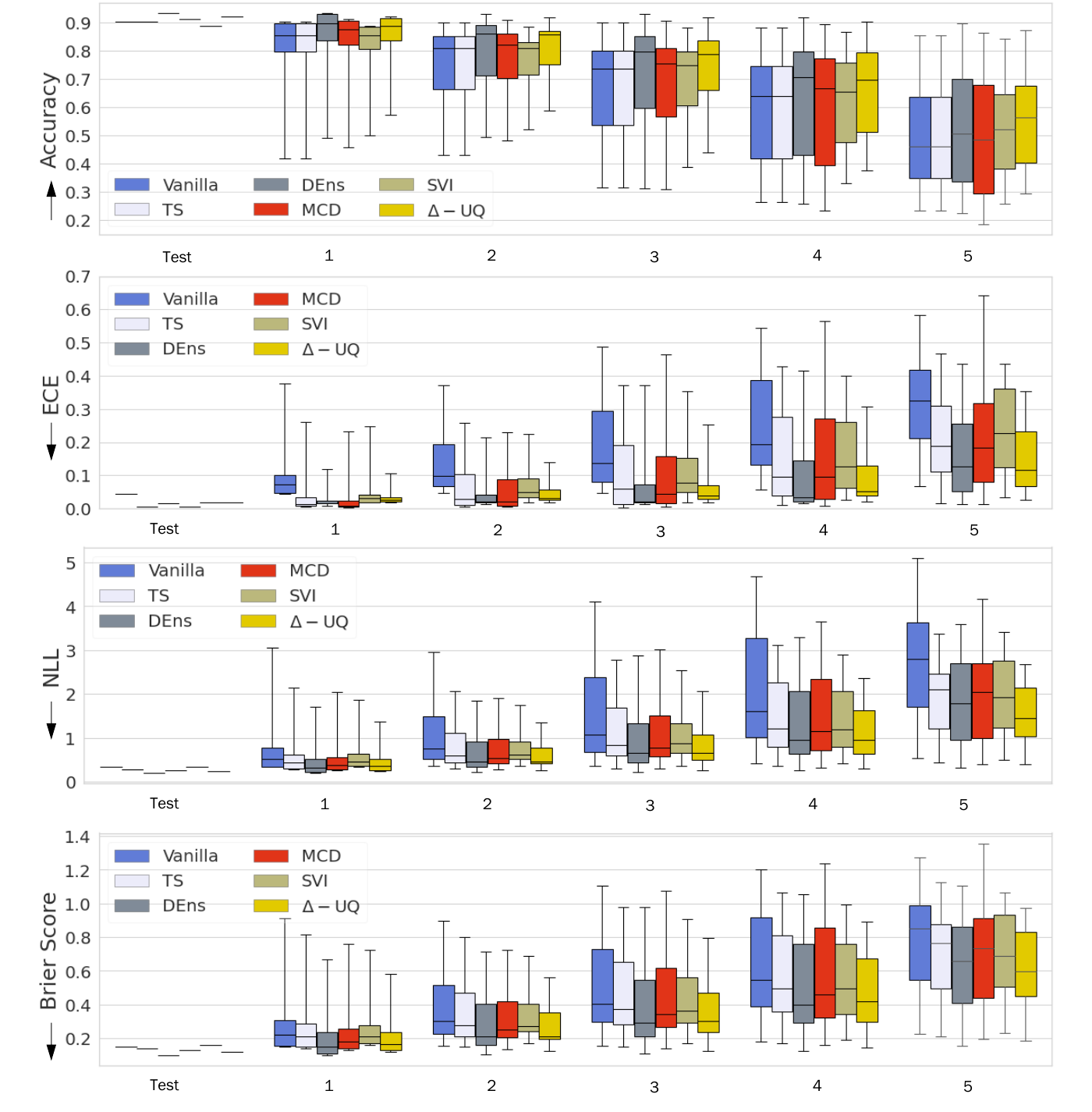}
    \caption{Calibration experiments using CIFAR-10C with a ResNet-20 model. We report average metrics for model accuracy and calibration across 5 random seeds and corruption severity levels. Note, we calibrate our predictions as before using a scaling strategy as: $\bar{\boldsymbol\mu} = \boldsymbol\mu(1 - \sigma)$. We compare against standard baselines obtained from \cite{krishnan2020improving, Ovadia2019}.}
    \label{fig:cifar10C}
\end{figure}

\paragraph{CIFAR-10C/ResNet-20} We perform detailed analysis of calibration and accuracy on CIFAR-10 and its corrupted variants CIFAR-10C \cite{hendrycks2018benchmarking} using the experimental protocol followed by \cite{Ovadia2019, krishnan2020improving}, where we use a ResNet-20 \cite{he2016deep} and report the calibration scores across all 5 corruption levels and the validation set -- the calibration metrics are reported by averaging the performance across 5 random seeds of the model. We report the average accuracy for each corruption level in table \ref{tab:acc-cifar-c}, and display the calibration metrics -- ECE, NLL and Brier Score in figure \ref{fig:cifar10C}. In both the accuracy and calibration metrics we find that $\Delta-$UQ outperforms all the comparable baselines, including Deep Ensembles, though using only a single model. 

\begin{table}[!htb]
\centering
\caption{Accuracy of ResNet-20 Model on CIFAR10 validation and its distribution shifted variants. }
\label{tab:acc-cifar-c}
\begin{tabular}{cccccccc}
\toprule
\textbf{Method} & \multicolumn{7}{c}{\textbf{CIFAR10-C Dist. Shift Variants (ResNet-20)}}                      \\
\cline{2-8}
                & \textbf{val} & \textbf{1} & \textbf{2} & \textbf{3} & \textbf{4} & \textbf{5} & \textbf{Avg.} \\
\toprule
Vanilla         & 90.5         & 81.8       & 75.1       & 68.3       & 60.6       & 49.1       & 69.8          \\
DEns            & 93.4         & 85.9       & 79.8       & 73.1       & 65         & 52.4       & 72.9 \\
MC Dropout      & 91           & 83.7       & 77.5       & 70.1       & 61.5       & 49.4       & 70.2          \\
SVI             & 88.6         & 82.3       & 76.9       & 70.8       & 63.1       & 52.6       & 70.6          \\
\midrule
$\Delta-$UQ & 92.3 & 85.8 & 80.9 & 75.2 & 67.8 & 56.2 & 74.25 \\
\bottomrule
\end{tabular}
\end{table}

\paragraph{Ablation studies.} We mainly perform ablation on the inference part of $\Delta-$UQ -- we try to quantify the variability in performance when anchoring is not used, and when uncertainty is not used in terms of the calibration metrics for the ImageNet-C/ResNet-50 experiment considered in the main paper. We consider two main ablations of the main model as stated next. Note, in all three cases the only difference is the procedure for inference. The training procedure is kept fixed across all models, and we use the same ResNet-50 model to perform these ablation studies reported in table \ref{tab:ablation}.
\begin{itemize}
    \item \textbf{Na\"ive}: We consider the case where a model trained with anchoring as usual, is used during inference \emph{without anchoring}, i.e., instead of passing $\{\mathrm{c}, \mathrm{x-c}\}$ as before for an anchor $\mathrm{c}$, we pass $\{0,\mathrm{x}\}$, as this behaves as a naive model that does not have the benefit of obtaining uncertainties or ensembling like behavior as $\Delta-UQ$. 
    \item \textbf{Ensemble mean}: Next, we consider the version where anchoring is done during inference and compute the mean of predictions from different anchors as before, but we do not use the uncertainties obtained -- i.e., the final prediction is simply the mean of the predictions obtained with different anchors. 
    \item $\Delta-$UQ: This is our final model that takes the mean and scales it by the uncertainties during inference. 
\end{itemize}

We observe from the results in table \ref{tab:ablation} first that simply training with anchoring shows benefits in model performance even if anchoring is not used during inference, as seen in improvement in the calibration performance of the na\"ive model (shown as $\{0,\mathrm{x}\}$ in the table) over the vanilla model. Next, we see that using anchoring to compute the mean improves performance further as seen in the next column (shown as $\boldsymbol \mu$), and finally factoring in the uncertainties performs the best. Even the ablated versions perform competitively compared to some of the other uncertainty baselines, indicating the effectiveness of \name. 

\begin{table}[!t]
\caption{\small{\textbf{Calibration Comparison With Ablation:} We study how different ablations of our model perform on the calibration task.}}
\label{tab:ablation}
\small{\begin{tblr}{
colspec ={X[1] X[1.8] X[1] X[1] X[1] X[1] X[1] X[1.2] X[1.2] X[1]},
hlines = {black,0pt},
vlines = {black,0.5pt},
cell{1}{1} = {r=2, c=1}{c,headers,fg=white},
cell{1}{2} = {r=2, c=1}{c,headers,fg=white},
cell{1}{3} = {r=2, c=1}{c,headers,fg=white},
cell{1}{4} = {r=2, c=1}{c,headers,fg=white},
cell{1}{5} = {r=2, c=1}{c,headers,fg=white},
cell{1}{6} = {r=2, c=1}{c,headers,fg=white},
cell{1}{7} = {r=2, c=1}{c,headers,fg=white},
cell{1}{8} = {r=2, c=1}{c},
cell{1}{9} = {r=2, c=1}{c},
cell{1}{10} = {r=2, c=1}{c},
row{3-14}  = {white, c},
cell{3}{1} = {r=4,c=1}{c,cellcol},
cell{7}{1} = {r=4,c=1}{c,cellcol},
cell{11}{1} = {r=4,c=1}{c,cellcol},
cell{14}{10} = {white}, cell{14}{9} = {white}, cell{14}{5} = {white},
cell{13}{10} = {white}, cell{13}{9} = {white}, cell{13}{5} = {white},
cell{12}{10} = {white}, cell{12}{5} = {white},
cell{11}{10} = {white}, cell{11}{5} = {white},
cell{10}{10} = {white}, cell{10}{9} = {white},
cell{9}{10} = {white}, cell{9}{5} = {white},
cell{8}{10} = {white}, cell{8}{5} = {white},
cell{7}{10} = {white}, cell{7}{5} = {white},
cell{6}{10} = {white}, cell{6}{9} = {white},
cell{5}{10} = {white}, cell{5}{9} = {white},
cell{4}{10} = {white}, cell{4}{9} = {white},
cell{3}{10} = {white}, cell{3}{9} = {white},
colsep=3pt, rowsep=0.5pt,hspan=minimal
                }
\hline
{Metric}            &                   &Vanilla & \small{Temp Scaling} & {DEns}& {MCD} & {SVI-AvUC} & {\color{blue}$\{0,\mathrm{x}\}$} & {\color{blue}($\boldsymbol\mu$)} & {\color{blue}\name}
\\
&&&&&&&&&\\ \hline
ECE $\downarrow$    & lower quartile    & 0.124 & 0.096 & 0.050 & 0.078 & 0.032 & 0.077 & 0.085 & 0.022\\
                    & median            & 0.174 & 0.139 & 0.090 & 0.134 & 0.045 & 0.117 & 0.112 & 0.038\\
                    & mean              & 0.194 & 0.160 & 0.088 & 0.153 & 0.054 & 0.130 & 0.110 & 0.044\\
                    & upper quartile    & 0.274 & 0.236 & 0.126 & 0.219 & 0.070 & 0.193 & 0.125 & 0.063\\ \hline
                    
 NLL $\downarrow$   & lower quartile    & 4.635 & 4.53 & 4.035 & 4.699 & 4.164 & 4.011 & 4.072 & 4.014\\
                    & median            & 5.115 & 4.993 & 4.624 & 5.093 & 4.823 & 4.818 & 4.679 & 4.617\\
                    & mean              & 5.234 & 5.091 & 4.604 & 5.553 & 4.707 & 4.832 & 4.516 & 4.352\\
                    & upper quartile    & 6.292 & 6.165 & 5.893 & 6.522 & 5.778 & 5.925 & 5.124 & 4.987\\ \hline
                    
 Brier $\downarrow$ &lower quartile     & 0.941 & 0.926 & 0.877 & 0.933 & 0.883 & 0.882 & 0.887 & 0.868\\
                    & median            & 0.987 & 0.970 & 0.922 & 0.967 & 0.935 & 0.944 & 0.940 & 0.925\\
                    & mean              & 0.964 & 0.945 & 0.888 & 0.961 & 0.900 & 0.926 & 0.903 & 0.887\\
                    & upper quartile    & 1.052 & 1.027 & 0.989 & 1.025 & 0.985 & 1.026 & 0.972 & 0.949\\\hline
\end{tblr}
}
\vspace{-10pt}
\end{table}

%% file: supp_gan_expt.tex
\section{Details on GAN-based Optimization Experiment}
In this experiment, we evaluated the utility of the proposed uncertainty estimator in guiding sequential optimization in the latent space of a pre-trained GAN network. We begin by assuming access to a generative model $\mathrm{G}(\mathrm{z})$, which maps a latent noise vector $\mathrm{z}$ onto a realization on the training image manifold. Denoting the latent space as $\mathcal{Z}$, our goal is to maximize a scalar function defined for an image, i.e., $f(\mathrm{x})$ by performing optimization in the latent space.
\begin{equation}
  \arg \max_{\mathrm{z}\in \mathcal{Z}} f(\mathrm{G}(\mathrm{z})).
\end{equation}In our experiment, we used a GAN trained on MNIST hand-written images and defined the thickness function (total number of non-zero pixels in an image) for optimization. The dimensionality of the noise latent space was set to $100$. Similar to our design optimization experiments with synthetic data, we started with an initial random sample (uniform random in the latent space) of $25$, synthesized the corresponding images using the generator and computed the thickness function for each of them. We performed optimization for $50$ steps (with $1$ sample in each round) and evaluated the maximum thickness achieved as the metric of choice (the global optimum is not known). We repeated the experiment across $5$ random seeds and $5$ independent trials for each seed. The results in Figure 7 (main paper) illustrate the maximum thickness obtained using different uncertainty estimators across $25$ experiments. We find that our approach consistently produced the highest function value and outperformed other approaches. As expected, DEns performed the second best, followed by MCD. Interestingly, with BNNs, the variational inferencing technique is known to lead to underfitting and we find out, despite achieving reasonably higher function values, the reconstructions were of significantly poorer quality (off the image manifold). 

%% file: supp_design_opt.tex
\section{Benchmark Functions}
\begin{figure*}[htb!]
    \centering
    \includegraphics[width = 0.99\textwidth]{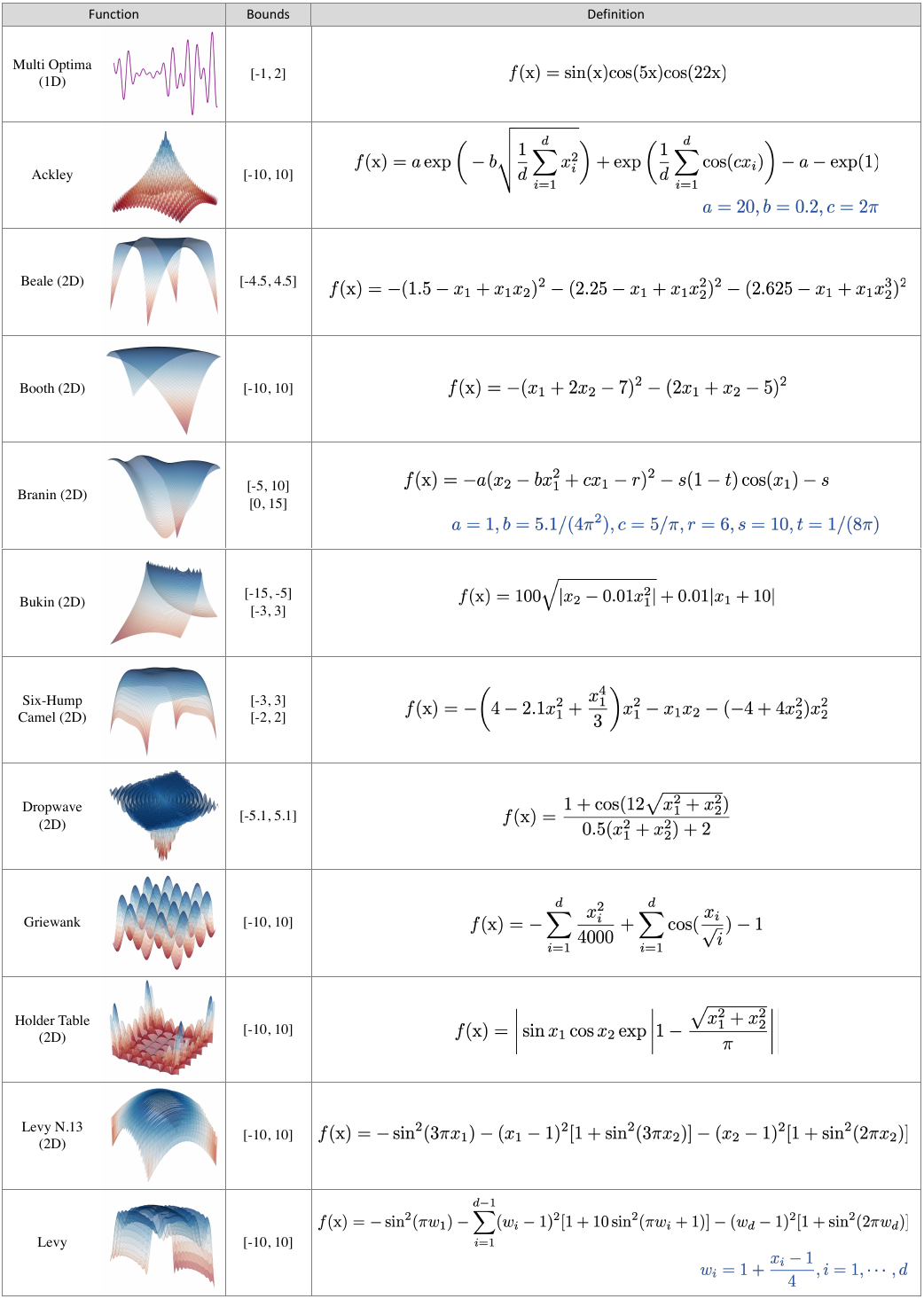}
    \caption{Benchmark functions used in this paper to evaluate sequential optimization}
    \label{fig:functions}
\end{figure*}

Figure \ref{fig:functions} illustrates the different benchmark functions for evaluating the proposed approach in black-box optimization. In addition to the functions listed, we also considered the Hartmann functions, in dimensions $3$ and $6$ respectively, defined as follows.

\begin{align}
\nonumber \textbf{Hartmann3: } &f(\mathrm{x}) = \sum_{i=1}^4 \alpha_i \exp \bigg(-\sum_{j=1}^3 A_{ij}(x_j - P_{ij})^2\bigg), \text{ where} \\
&\nonumber \alpha = (1.0,1.2,3.0,3.2)^{\top} \\
&
  \mathbf{A} =
  \begin{pmatrix}
    3.0 & 10 & 30\\
    0.1 & 10 & 35\\
    3.0 & 10 & 30 \\
    0.1 & 10 & 35
  \end{pmatrix}, \quad
  \mathbf{P} = 10^{-4}
  \begin{pmatrix}
    3689 & 1170 & 2673\\
    4699 & 4387 & 7470\\
    1091 & 8732 & 5547 \\
    381 & 5743 & 8828
  \end{pmatrix}
\end{align}

\begin{align}
\nonumber \textbf{Hartmann6: } &f(\mathrm{x}) = \sum_{i=1}^4 \alpha_i \exp \bigg(-\sum_{j=1}^6 A_{ij}(x_j - P_{ij})^2\bigg), \text{ where} \\
&\nonumber \alpha = (1.0,1.2,3.0,3.2)^{\top} \\
&
  \nonumber \mathbf{A} =
  \begin{pmatrix}
    10 & 3 & 17 &3.50 &1.7 & 8\\
    0.05 & 10 & 17 & 0.1 & 8 & 4\\
    3.0 & 3.5 & 1.7 & 10 & 17 & 8 \\
   17 & 8 & 0.05 & 10 & 0.1 & 14
  \end{pmatrix}, \\
  &\mathbf{P} = 10^{-4}
  \begin{pmatrix}
    1312 & 1696 & 5569 & 124 & 8283 & 5886\\
    2329 & 4135 & 8307 & 3736 & 1004 & 9991\\
    2348 & 1451 & 3511 & 2883 & 3047 & 6650 \\
    4047 & 8828 & 8732 & 5743 & 1091 & 381
  \end{pmatrix}
\end{align}

\begin{figure}[htb!]
    \centering
    \includegraphics[width = 0.99\linewidth]{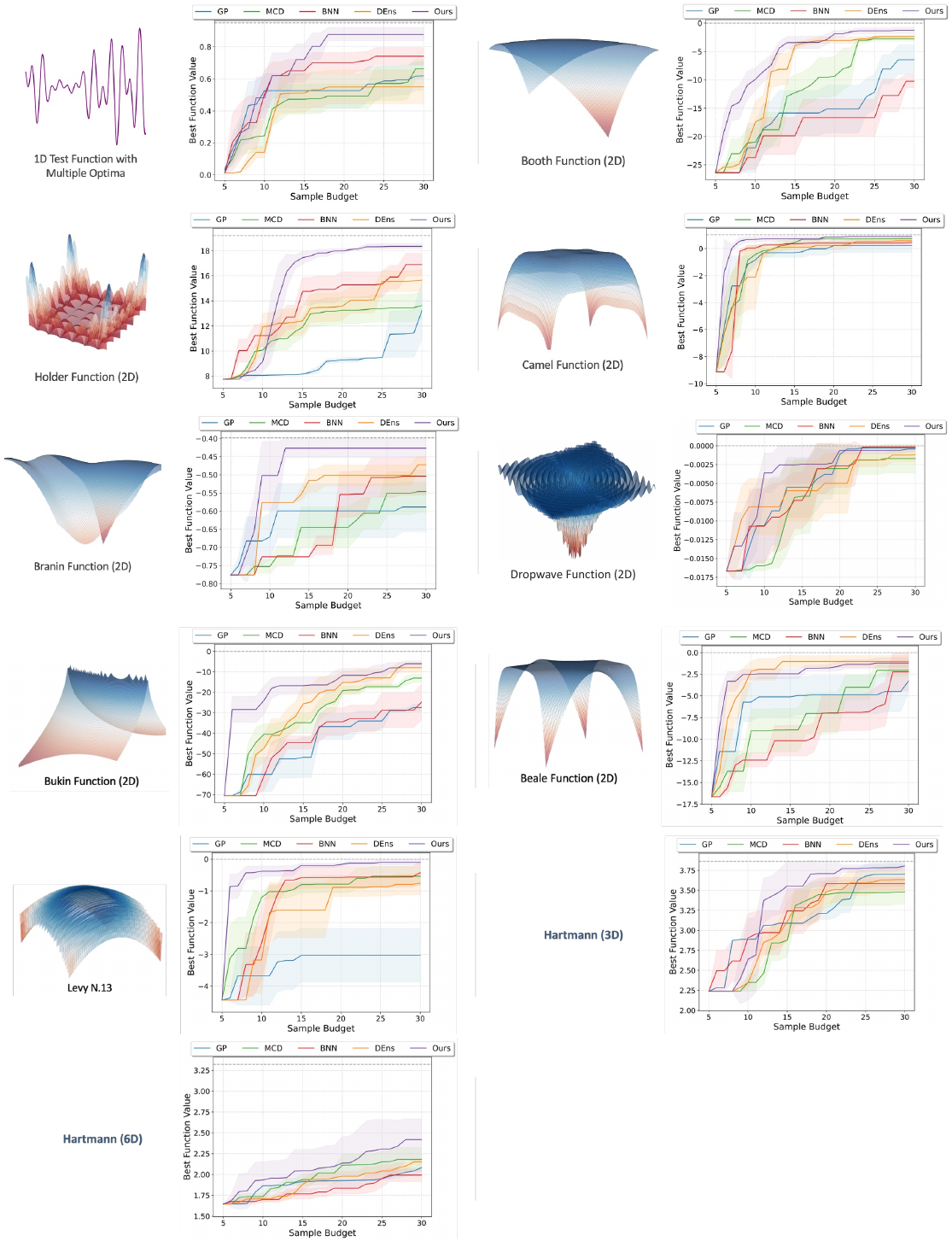}
    \caption{Convergence curves for each of the benchmark functions used in our evaluations.}
    \label{fig:desopt}
\end{figure}

\section{Detailed Results for Sequential Optimization}
At the core of AI-powered applications in science and engineering lies the need to perform design optimization for maximizing a chosen target objective, and to enable automated exploration in high-dimensional parameter spaces. When $f$ is Lipschitz continuous, i.e., $\|f(\mathrm{x}) - f(\mathrm{x}^{\prime})\| \leq c \|\mathrm{x} - \mathrm{x}^{\prime}\|$, and its first- and second-order information are accessible, this can be solved using first-order optimization methods such as stochastic gradient descent (SGD)~\cite{bottou2012stochastic} or second-order methods such as L-BFGS~\cite{liu1989limited}. However, in practice, while $f$ can be explicitly evaluated for any $\mathrm{x}$, its first- and second-order information are unknown, thus making such an optimization very challenging. Commonly referred to as \textit{black-box optimization}~\cite{audet2017derivative}, this formulation is adopted in applications ranging from drug design~\cite{schneider2020rethinking} to additive manufacturing~\cite{wang2020machine} and optimizing financial investments~\cite{gonzalvez2019financial} to hyper-parameter tuning in neural networks~\cite{ren2021comprehensive}. 
\begin{equation}
    \mathrm{x}^* = arg \max_{\mathrm{x} \in \mathcal{D}} \mathrm{f}(\mathrm{x}).
    \label{eq:basic}
\end{equation}

In each step of this optimization, we approximate the function $f$ using the samples observed so far to obtain the surrogate $\hat{f}$. Assuming that our goal is to maximize $f$, one can acquire more samples in regimes of $\mathcal{D}$ where the mean estimate from the surrogate is high (\textit{exploitation}) or the uncertainty is large (\textit{exploration}). In order to balance between these two objectives and to guide the progressive search for the optima, BO utilizes an appropriate \textit{acquisition} function~\cite{snoek2012practical}. In this study, we use the popular expected improvement (EI) score to perform candidate selection.
\begin{align}
\nonumber
\text{aq}_{\text{EI}} & \coloneqq 
\begin{cases}
  (\mu(\mathrm{x}) - f(\mathrm{x}^+) - \xi) \Phi(\mathrm{Z}) + \sigma(\mathrm{x}) \phi(\mathrm{Z}) & \text{if $\sigma(\mathrm{x})>0$} \\
  0 & \text{if $\sigma(\mathrm{x})=0$}
\end{cases} \\
&\text{where } \mathrm{Z} = \frac{(\mu(\mathrm{x}) - f(\mathrm{x}^+) - \xi)}{\sigma(\mathrm{x})}.
    \label{eq:ei}
\end{align}Here, $\mu(\mathrm{x})$ and $\sigma(\mathrm{x})$ are the mean and uncertainty estimates from the surrogate $\hat{f}$ for any sample, and $f(\mathrm{x}^+)$ is the best known function value so far during any iteration of the optimization. Further, $\Phi(.)$ and $\phi(.)$ denote the cumulative distribution and probability density functions corresponding to the normal distribution. Finally, the hyper-parameter $\xi$ controls the exploration-exploitation trade-off. Figure \ref{fig:desopt} shows the convergence curves for each of the black-box functions.

%% file: main.bbl
\begin{thebibliography}{39}
\providecommand{\natexlab}[1]{#1}
\providecommand{\url}[1]{\texttt{#1}}
\expandafter\ifx\csname urlstyle\endcsname\relax
  \providecommand{\doi}[1]{doi: #1}\else
  \providecommand{\doi}{doi: \begingroup \urlstyle{rm}\Url}\fi

\bibitem[Wilson and Izmailov(2020)]{wilson2020bayesian}
Andrew~G Wilson and Pavel Izmailov.
\newblock Bayesian deep learning and a probabilistic perspective of
  generalization.
\newblock \emph{Advances in neural information processing systems},
  33:\penalty0 4697--4708, 2020.

\bibitem[He et~al.(2020)He, Lakshminarayanan, and Teh]{he2020bayesian}
Bobby He, Balaji Lakshminarayanan, and Yee~Whye Teh.
\newblock Bayesian deep ensembles via the neural tangent kernel.
\newblock \emph{Advances in Neural Information Processing Systems},
  33:\penalty0 1010--1022, 2020.

\bibitem[Neal(2012)]{neal2012bayesian}
Radford~M Neal.
\newblock \emph{Bayesian learning for neural networks}, volume 118.
\newblock Springer Science \& Business Media, 2012.

\bibitem[Blundell et~al.(2015)Blundell, Cornebise, Kavukcuoglu, and
  Wierstra]{blundell2015weight}
Charles Blundell, Julien Cornebise, Koray Kavukcuoglu, and Daan Wierstra.
\newblock Weight uncertainty in neural network.
\newblock In \emph{International Conference on Machine Learning}, pages
  1613--1622. PMLR, 2015.

\bibitem[Gal and Ghahramani(2016)]{gal2016dropout}
Yarin Gal and Zoubin Ghahramani.
\newblock Dropout as a bayesian approximation: Representing model uncertainty
  in deep learning.
\newblock In \emph{international conference on machine learning}, pages
  1050--1059. PMLR, 2016.

\bibitem[Lakshminarayanan et~al.(2017)Lakshminarayanan, Pritzel, and
  Blundell]{lakshminarayanan2017simple}
Balaji Lakshminarayanan, Alexander Pritzel, and Charles Blundell.
\newblock Simple and scalable predictive uncertainty estimation using deep
  ensembles.
\newblock \emph{Advances in neural information processing systems}, 30, 2017.

\bibitem[Ovadia et~al.(2019)Ovadia, Fertig, Ren, Nado, Sculley, Nowozin,
  Dillon, Lakshminarayanan, and Snoek]{Ovadia2019}
Yaniv Ovadia, Emily Fertig, Jie Ren, Zachary Nado, David Sculley, Sebastian
  Nowozin, Joshua Dillon, Balaji Lakshminarayanan, and Jasper Snoek.
\newblock Can you trust your model's uncertainty? evaluating predictive
  uncertainty under dataset shift.
\newblock \emph{Advances in neural information processing systems}, 32, 2019.

\bibitem[Van~Amersfoort et~al.(2020)Van~Amersfoort, Smith, Teh, and
  Gal]{van2020uncertainty}
Joost Van~Amersfoort, Lewis Smith, Yee~Whye Teh, and Yarin Gal.
\newblock Uncertainty estimation using a single deep deterministic neural
  network.
\newblock In \emph{International Conference on Machine Learning}, pages
  9690--9700. PMLR, 2020.

\bibitem[Jain et~al.(2021)Jain, Lahlou, Nekoei, Butoi, Bertin, Rector-Brooks,
  Korablyov, and Bengio]{jain2021deup}
Moksh Jain, Salem Lahlou, Hadi Nekoei, Victor Butoi, Paul Bertin, Jarrid
  Rector-Brooks, Maksym Korablyov, and Yoshua Bengio.
\newblock Deup: Direct epistemic uncertainty prediction.
\newblock \emph{arXiv preprint arXiv:2102.08501}, 2021.

\bibitem[Jacot et~al.(2018)Jacot, Gabriel, and Hongler]{jacot2018neural}
Arthur Jacot, Franck Gabriel, and Cl{\'e}ment Hongler.
\newblock Neural tangent kernel: Convergence and generalization in neural
  networks.
\newblock \emph{Advances in neural information processing systems}, 31, 2018.

\bibitem[Tancik et~al.(2020)Tancik, Srinivasan, Mildenhall, Fridovich-Keil,
  Raghavan, Singhal, Ramamoorthi, Barron, and Ng]{tancik2020fourier}
Matthew Tancik, Pratul Srinivasan, Ben Mildenhall, Sara Fridovich-Keil, Nithin
  Raghavan, Utkarsh Singhal, Ravi Ramamoorthi, Jonathan Barron, and Ren Ng.
\newblock Fourier features let networks learn high frequency functions in low
  dimensional domains.
\newblock \emph{Advances in Neural Information Processing Systems},
  33:\penalty0 7537--7547, 2020.

\bibitem[Graves(2011)]{graves2011practical}
Alex Graves.
\newblock Practical variational inference for neural networks.
\newblock In \emph{Advances in neural information processing systems}, pages
  2348--2356. Citeseer, 2011.

\bibitem[Welling and Teh(2011)]{welling2011bayesian}
Max Welling and Yee~W Teh.
\newblock Bayesian learning via stochastic gradient langevin dynamics.
\newblock In \emph{Proceedings of the 28th international conference on machine
  learning (ICML-11)}, pages 681--688. Citeseer, 2011.

\bibitem[Fort et~al.(2019)Fort, Hu, and Lakshminarayanan]{fort2019deep}
Stanislav Fort, Huiyi Hu, and Balaji Lakshminarayanan.
\newblock Deep ensembles: A loss landscape perspective.
\newblock \emph{arXiv preprint arXiv:1912.02757}, 2019.

\bibitem[Arora et~al.(2019)Arora, Du, Hu, Li, and Wang]{arora2019fine}
Sanjeev Arora, Simon Du, Wei Hu, Zhiyuan Li, and Ruosong Wang.
\newblock Fine-grained analysis of optimization and generalization for
  overparameterized two-layer neural networks.
\newblock In \emph{International Conference on Machine Learning}, pages
  322--332. PMLR, 2019.

\bibitem[Bietti and Mairal(2019)]{bietti2019inductive}
Alberto Bietti and Julien Mairal.
\newblock On the inductive bias of neural tangent kernels.
\newblock \emph{Advances in Neural Information Processing Systems}, 32, 2019.

\bibitem[Lee et~al.(2019)Lee, Xiao, Schoenholz, Bahri, Novak, Sohl-Dickstein,
  and Pennington]{lee2019wide}
Jaehoon Lee, Lechao Xiao, Samuel Schoenholz, Yasaman Bahri, Roman Novak, Jascha
  Sohl-Dickstein, and Jeffrey Pennington.
\newblock Wide neural networks of any depth evolve as linear models under
  gradient descent.
\newblock \emph{Advances in neural information processing systems}, 32, 2019.

\bibitem[Lee et~al.(2018)Lee, Sohl-dickstein, Pennington, Novak, Schoenholz,
  and Bahri]{lee2018deep}
Jaehoon Lee, Jascha Sohl-dickstein, Jeffrey Pennington, Roman Novak, Sam
  Schoenholz, and Yasaman Bahri.
\newblock Deep neural networks as gaussian processes.
\newblock In \emph{International Conference on Learning Representations}, 2018.
\newblock URL \url{https://openreview.net/forum?id=B1EA-M-0Z}.

\bibitem[de~G.~Matthews et~al.(2018)de~G.~Matthews, Hron, Rowland, Turner, and
  Ghahramani]{gMatthews2018gaussian}
Alexander~G. de~G.~Matthews, Jiri Hron, Mark Rowland, Richard~E. Turner, and
  Zoubin Ghahramani.
\newblock Gaussian process behaviour in wide deep neural networks.
\newblock In \emph{International Conference on Learning Representations}, 2018.
\newblock URL \url{https://openreview.net/forum?id=H1-nGgWC-}.

\bibitem[Novak et~al.(2019)Novak, Xiao, Bahri, Lee, Yang, Abolafia, Pennington,
  and Sohl-dickstein]{novak2019bayesian}
Roman Novak, Lechao Xiao, Yasaman Bahri, Jaehoon Lee, Greg Yang, Daniel~A.
  Abolafia, Jeffrey Pennington, and Jascha Sohl-dickstein.
\newblock Bayesian deep convolutional networks with many channels are gaussian
  processes.
\newblock In \emph{International Conference on Learning Representations}, 2019.
\newblock URL \url{https://openreview.net/forum?id=B1g30j0qF7}.

\bibitem[Woodbury(1950)]{woodbury1950inverting}
Max~A Woodbury.
\newblock \emph{Inverting modified matrices}.
\newblock Statistical Research Group, 1950.

\bibitem[Russakovsky et~al.(2015)Russakovsky, Deng, Su, Krause, Satheesh, Ma,
  Huang, Karpathy, Khosla, Bernstein, et~al.]{russakovsky2015imagenet}
Olga Russakovsky, Jia Deng, Hao Su, Jonathan Krause, Sanjeev Satheesh, Sean Ma,
  Zhiheng Huang, Andrej Karpathy, Aditya Khosla, Michael Bernstein, et~al.
\newblock Imagenet large scale visual recognition challenge.
\newblock \emph{International journal of computer vision}, 115\penalty0
  (3):\penalty0 211--252, 2015.

\bibitem[He et~al.(2016)He, Zhang, Ren, and Sun]{he2016deep}
Kaiming He, Xiangyu Zhang, Shaoqing Ren, and Jian Sun.
\newblock Deep residual learning for image recognition.
\newblock In \emph{Proceedings of the IEEE conference on computer vision and
  pattern recognition}, pages 770--778, 2016.

\bibitem[Guo et~al.(2017)Guo, Pleiss, Sun, and Weinberger]{guo2017calibration}
Chuan Guo, Geoff Pleiss, Yu~Sun, and Kilian~Q Weinberger.
\newblock On calibration of modern neural networks.
\newblock In \emph{International Conference on Machine Learning}, pages
  1321--1330. PMLR, 2017.

\bibitem[Krishnan and Tickoo(2020)]{krishnan2020improving}
Ranganath Krishnan and Omesh Tickoo.
\newblock Improving model calibration with accuracy versus uncertainty
  optimization.
\newblock In \emph{Advances in Neural Information Processing Systems},
  volume~33, pages 18237--18248, 2020.

\bibitem[Hendrycks and Dietterich(2019)]{hendrycks2018benchmarking}
Dan Hendrycks and Thomas Dietterich.
\newblock Benchmarking neural network robustness to common corruptions and
  perturbations.
\newblock In \emph{International Conference on Learning Representations}, 2019.
\newblock URL \url{https://openreview.net/forum?id=HJz6tiCqYm}.

\bibitem[Liu et~al.(2020)Liu, Wang, Owens, and Li]{energyood}
Weitang Liu, Xiaoyun Wang, John Owens, and Yixuan Li.
\newblock Energy-based out-of-distribution detection.
\newblock \emph{Advances in Neural Information Processing Systems},
  33:\penalty0 21464--21475, 2020.

\bibitem[Kailkhura et~al.(2018)Kailkhura, Thiagarajan, Rastogi, Varshney, and
  Bremer]{kailkhura2018spectral}
Bhavya Kailkhura, Jayaraman~J Thiagarajan, Charvi Rastogi, Pramod~K Varshney,
  and Peer-Timo Bremer.
\newblock A spectral approach for the design of experiments: Design, analysis
  and algorithms.
\newblock \emph{The Journal of Machine Learning Research}, 19\penalty0
  (1):\penalty0 1214--1259, 2018.

\bibitem[Shahriari et~al.(2015)Shahriari, Swersky, Wang, Adams, and
  De~Freitas]{shahriari2015taking}
Bobak Shahriari, Kevin Swersky, Ziyu Wang, Ryan~P Adams, and Nando De~Freitas.
\newblock Taking the human out of the loop: A review of bayesian optimization.
\newblock \emph{Proceedings of the IEEE}, 104\penalty0 (1):\penalty0 148--175,
  2015.

\bibitem[Snoek et~al.(2012)Snoek, Larochelle, and Adams]{snoek2012practical}
Jasper Snoek, Hugo Larochelle, and Ryan~P Adams.
\newblock Practical bayesian optimization of machine learning algorithms.
\newblock \emph{Advances in neural information processing systems}, 25, 2012.

\bibitem[Sun et~al.(2017)Sun, Chen, and Carin]{sun2017learning}
Shengyang Sun, Changyou Chen, and Lawrence Carin.
\newblock Learning structured weight uncertainty in bayesian neural networks.
\newblock In \emph{Artificial Intelligence and Statistics}, pages 1283--1292.
  PMLR, 2017.

\bibitem[Nado et~al.(2021)Nado, Band, Collier, Djolonga, Dusenberry, Farquhar,
  Filos, Havasi, Jenatton, Jerfel, Liu, Mariet, Nixon, Padhy, Ren, Rudner, Wen,
  Wenzel, Murphy, Sculley, Lakshminarayanan, Snoek, Gal, and
  Tran]{nado2021uncertainty}
Zachary Nado, Neil Band, Mark Collier, Josip Djolonga, Michael Dusenberry,
  Sebastian Farquhar, Angelos Filos, Marton Havasi, Rodolphe Jenatton, Ghassen
  Jerfel, Jeremiah Liu, Zelda Mariet, Jeremy Nixon, Shreyas Padhy, Jie Ren, Tim
  Rudner, Yeming Wen, Florian Wenzel, Kevin Murphy, D.~Sculley, Balaji
  Lakshminarayanan, Jasper Snoek, Yarin Gal, and Dustin Tran.
\newblock {Uncertainty Baselines}: Benchmarks for uncertainty \& robustness in
  deep learning.
\newblock \emph{arXiv preprint arXiv:2106.04015}, 2021.

\bibitem[Bottou(2012)]{bottou2012stochastic}
L{\'e}on Bottou.
\newblock Stochastic gradient descent tricks.
\newblock In \emph{Neural networks: Tricks of the trade}, pages 421--436.
  Springer, 2012.

\bibitem[Liu and Nocedal(1989)]{liu1989limited}
Dong~C Liu and Jorge Nocedal.
\newblock On the limited memory bfgs method for large scale optimization.
\newblock \emph{Mathematical programming}, 45\penalty0 (1):\penalty0 503--528,
  1989.

\bibitem[Audet and Hare(2017)]{audet2017derivative}
Charles Audet and Warren Hare.
\newblock \emph{Derivative-free and blackbox optimization}, volume~2.
\newblock Springer, 2017.

\bibitem[Schneider et~al.(2020)Schneider, Walters, Plowright, Sieroka,
  Listgarten, Goodnow, Fisher, Jansen, Duca, Rush,
  et~al.]{schneider2020rethinking}
Petra Schneider, W~Patrick Walters, Alleyn~T Plowright, Norman Sieroka,
  Jennifer Listgarten, Robert~A Goodnow, Jasmin Fisher, Johanna~M Jansen,
  Jos{\'e}~S Duca, Thomas~S Rush, et~al.
\newblock Rethinking drug design in the artificial intelligence era.
\newblock \emph{Nature Reviews Drug Discovery}, 19\penalty0 (5):\penalty0
  353--364, 2020.

\bibitem[Wang et~al.(2020)Wang, Tan, Tor, and Lim]{wang2020machine}
Chengcheng Wang, XP~Tan, SB~Tor, and CS~Lim.
\newblock Machine learning in additive manufacturing: State-of-the-art and
  perspectives.
\newblock \emph{Additive Manufacturing}, 36:\penalty0 101538, 2020.

\bibitem[Gonzalvez et~al.(2019)Gonzalvez, Lezmi, Roncalli, and
  Xu]{gonzalvez2019financial}
Joan Gonzalvez, Edmond Lezmi, Thierry Roncalli, and Jiali Xu.
\newblock Financial applications of gaussian processes and bayesian
  optimization.
\newblock \emph{arXiv preprint arXiv:1903.04841}, 2019.

\bibitem[Ren et~al.(2021)Ren, Xiao, Chang, Huang, Li, Chen, and
  Wang]{ren2021comprehensive}
Pengzhen Ren, Yun Xiao, Xiaojun Chang, Po-Yao Huang, Zhihui Li, Xiaojiang Chen,
  and Xin Wang.
\newblock A comprehensive survey of neural architecture search: Challenges and
  solutions.
\newblock \emph{ACM Computing Surveys (CSUR)}, 54\penalty0 (4):\penalty0 1--34,
  2021.

\end{thebibliography}
